\documentclass[sensors,article,accept,moreauthors,pdftex]{Definitions/mdpi} 

\usepackage{acronym}
\usepackage{siunitx}
\usepackage{commath}
\usepackage{bm}
\usepackage{stmaryrd}
\usepackage{dsfont}

\newcommand*\patchAmsMathEnvironmentForLineno[1]{%
  \expandafter\let\csname old#1\expandafter\endcsname\csname #1\endcsname
  \expandafter\let\csname oldend#1\expandafter\endcsname\csname end#1\endcsname
  \renewenvironment{#1}%
     {\linenomath\csname old#1\endcsname}%
     {\csname oldend#1\endcsname\endlinenomath}}%
\newcommand*\patchBothAmsMathEnvironmentsForLineno[1]{%
  \patchAmsMathEnvironmentForLineno{#1}%
  \patchAmsMathEnvironmentForLineno{#1*}}%
\AtBeginDocument{%
\patchBothAmsMathEnvironmentsForLineno{equation}%
\patchBothAmsMathEnvironmentsForLineno{align}%
\patchBothAmsMathEnvironmentsForLineno{flalign}%
\patchBothAmsMathEnvironmentsForLineno{alignat}%
\patchBothAmsMathEnvironmentsForLineno{gather}%
\patchBothAmsMathEnvironmentsForLineno{multline}%
}

\newcommand{\floor}[1]{\lfloor #1 \rfloor}

\newcommand{\expp}[1]{\exp\left(#1\right)}
\newcommand{\pr}[1]{\mathbb{P}\left(#1\right)}
\newcommand{\area}{\Delta a}

\renewcommand{\P}[1]{\mathbb{P}\left(#1\right)}
\MakeRobust{\overrightarrow}

\newif\ifhighlightReviews

\highlightReviewstrue

\acrodef{lidar}{Light Detection And Ranging}
\firstpage{1} 
\pubvolume{1}
\issuenum{1}
\articlenumber{0}
\pubyear{2021}
\copyrightyear{2021}
\externaleditor{Academic Editor: Ramon Barber 
 Firstname Lastname} 
\datereceived{} 
\dateaccepted{} 
\datepublished{} 
\hreflink{https://doi.org/}%
\Title{{A Novel Occupancy Mapping Framework for Risk-Aware Path Planning in Unstructured Environments}}

\TitleCitation{A Novel Occupancy Mapping Framework for Risk-Aware Path Planning in Unstructured Environments}

\Author{Johann Laconte
 $^{1,}$*, Abderrahim Kasmi $^{2}$, François Pomerleau $^{3}$, Roland Chapuis $^{1}$, Laurent Malaterre $^{1}$, \linebreak Christophe Debain $^{4}$ and Romuald Aufrère $^{1}$}
\AuthorNames{Johann Laconte, Abderrahim Kasmi, François Pomerleau, Roland Chapuis, Laurent Malaterre, Christophe Debain and Romuald Aufrère}

\AuthorCitation{Laconte, J.;
 Kasmi, A.; Pomerleau, F.; Chapuis, R.; Malaterre, L.; Debain, C.; Aufrère, R.}

\address{%
  $^{1}$ \quad 
  Université Clermont Auvergne, CNRS, Clermont Auvergne INP, Institut
Pascal, F-63000 Clermont-Ferrand, France; roland.chapuis@uca.fr (R.C.); laurent.malaterre@uca.fr (L.M.);  romuald.aufrere@uca.fr (R.A.)\\
$^{2}$ \quad Sherpa Engineering, R\&D Department, 333 Avenue Georges Clemenceau, 92000 Nanterre, France; abderrahim.kasmi@etu.uca.fr \\
$^{3}$ \quad Norlab, Université Laval, Québec, QC, Canada G1V 0A6;
 francois.pomerleau@ift.ulaval.ca\\
$^{4}$ \quad 
Université Clermont Auvergne, INRAE, UR TSCF, F-63178 Aubière, France; christophe.debain@inrae.fr}
 
\corres{Correspondence: johann.laconte@uca.fr}

\abstract{
	In the context of autonomous robots, one of the most important tasks is to prevent potential damage to the robot during navigation.
	For this purpose, it is often assumed that one must deal with known probabilistic obstacles, then compute the probability of collision with each obstacle.
	However, in complex scenarios or unstructured environments, it might be difficult to detect such obstacles.
	In these cases, a metric map is used, where each position stores the information of occupancy.
	The most common type of metric map is the Bayesian occupancy map.
	However, this type of map is not well suited for computing risk assessments for continuous paths due to its discrete nature.
	Hence, we introduce a novel type of map called the Lambda Field, which is specially designed for risk assessment.
	We first propose a way to compute such a map and the expectation of a generic risk over a path.
	Then, we demonstrate the benefits of our generic formulation with a use case defining the risk as the expected collision force over a path.
        Using this risk definition and the Lambda Field, we show that our framework is capable of doing classical path planning while having a physical-based metric.
	Furthermore, the Lambda Field gives a natural way to deal with unstructured environments, such as tall grass. %
	Where standard environment representations would always generate trajectories going around such obstacles, our framework allows the robot to go through the grass while being aware of the risk taken. %
}


\begin{document}

\section{Introduction} 
		Nowadays, autonomous robots are more and more visible in our lives.
		They start to prove themselves useful in a very broad spectrum of applications, from~autonomous driving to supporting humans in dangerous jobs such as  mining or search~and~rescue missions.
		One common aspect of every robot's tasks is the notion of safety; before taking any action, the~robots have to assess the associated risk of the action.

		To assess such a risk, robots need a way to represent and store the surrounding environment.
		In structured and controlled environments, such as warehouses, the~easiest solution is to provide the robot with a map of the environment, as well as the positions of every obstacle, robot, and operator.
                Storing such entities leads to the construction of semantic maps, where each obstacle is stored as an object (e.g., a wall, operator, or robot). 
		Under this representation, the~robot has to keep track of every moving obstacle while avoiding collisions with the environment.
		However, such a representation of the environment is not always available or easy to build from raw data in all situations.
		For example, it is impossible to perfectly describe the underlying environment of a snowy forest or a crowded park.
		There are indeed many unstructured obstacles in the first case that are not easily storable in such semantic maps, while in the second one, a lot of dynamic obstacles hinder the construction of a precise map.
		Clustering raw data from \ac{lidar} measurements, as~done by \citet{Fulgenzi2007}, for example, might not be possible for the aforementioned scenarios.

\begin{figure}[H]
				
				\includegraphics[width=.8\linewidth]{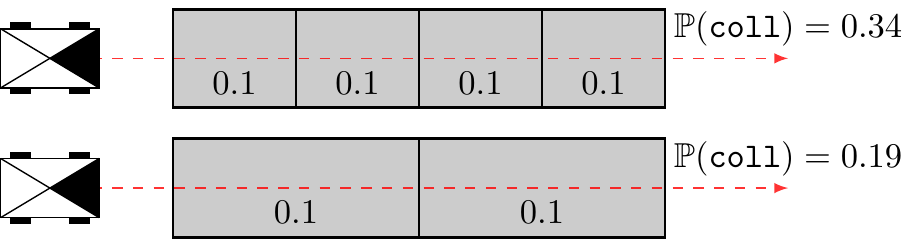}
				\caption{Example of collision assessment in an occupancy grid.
					The robots (black boxes with their front represented as a filled triangle) want to cross an environment by following the dashed red line.
					The collision probability is uniform for the whole environment ($0.1$).
					The discretization size greatly influences the probability of collision, with~the bottom scenario yielding a safer path even though the underlying environment is the same.
				}
				\label{fig:intro}
		\end{figure}

		When such high-level environmental representation is not available or possible, a~lower-level map is constructed, which called a metric map.
		Instead of storing features, the~metric map tessellates the environment into cells, where each one stores the information of occupancy.
		This kind of map has been heavily studied and used since the beginning of robotics.
		They were introduced by \citet{Elfes1989}, who proposed the concept of occupancy grids.
		Each cell stores the probability that the underlying environment is occupied and, hence, not traversable for the robot.
		This type of map is easy to construct and can be used to perform a great variety of tasks, such as Simultaneous Localization and Mapping (SLAM) or path planning.
		However, as~previously demonstrated by \citet{Heiden2017}, a~problem quickly arises when the robot wants to assess the probability of collision for a given path.
		Indeed, we are tempted to assess the probability of collision as the joint probability that every cell is free of obstacles.
		As an example, Figure \ref{fig:intro} shows a robot crossing an environment where the probability of occupancy is $0.1$ for each cell.
		Depending on the tessellation size, the~probability of collision can be $0.19$ or~$0.34$ for a tessellation half as small as the first one. 
		This behavior comes from the fact that the correlation between the occupancy of two positions of the environment is not null.
		Nevertheless, it is impossible to accurately estimate this correlation.
		The same problem arises when dealing with occupancy grids stored in quad-trees \citep{Kraetzschmar2004}.
		Indeed, the~robot could decide to cross a large high-probability cell instead of ten small low-probability~ones.

		Furthermore, the~probability of collision is not well suited to describing the risk in complex~situations.
		For example, crossing a part of the environment at $\SI[per-mode=symbol]{5}{\km\per\hour}$ is not as risky as crossing the same environment at $\SI[per-mode=symbol]{70}{\km\per\hour}$.
		The damages caused by a potential collision are far more consequential at a higher~speed.
	
		Under these considerations, we introduce the concept of the Lambda Field.
		 The Lambda Field is a representation of an environment that allows the computation of the probability of collision while being independent of the tessellation size.
		It also provides a natural way to assess more complex risks than the probability of collision.
                {Our framework consists of a novel occupancy-mapping technique, as well as a formulation for assessing risks on it, thus yielding a direct way for path-planning algorithms to work on these maps.}
                {Our key contributions are:}
		\begin{itemize}
                  \item {A novel type of map called the Lambda Field, which is specially designed to allow generic risk assessments and is better fitted for unstructured environments;}
                  \item {A mathematical formulation of risk assessment over a path with an application to path planning in tall grass; }
                  \item {\textls[-15]{A theoretical and experimental evaluation of the Bayesian occupancy grid, showing that such a framework can over-converge in the case of unstructured and sparse~obstacles.}}
		\end{itemize}

              {In this paper, we provide a revised and extended version of our previous work} \cite{Laconte2019}.
		We give an extended theory that takes into account the mass of the obstacles, allowing the robot to move in unstructured environments.
		Moreover, we improve the theory of \citet{Heiden2017} to take into account the robot size and prove that the Lambda Field is a generalization of their theory. 
		We also extend our results with tests in real-world conditions, in~both structured and unstructured environments, showing that Lambda Fields better map unstructured environments and allow behaviors that are impossible using classical occupancy~grids.

\section{Related~Work} \label{sec:relatedWorks}
	\textls[-5]{In order to perform path planning, the~first step is to construct a representation of the environment.
	In the context of unstructured or complex environments, a~semantic map is impossible to create and a lower-level representation, called a metric map, has to be used.
	This kind of map tessellates the environment into cells, where each one stores the information of occupancy.
	The idea of tessellating the sensed environment was originally proposed by \citet{Elfes1989}.
	Later, \citet{Coue2006} enhanced this idea by adding a Bayesian layer, which better handles uncertainty and noisy readings, increasing the robustness of the map.
	Many variations of the Bayesian occupancy filter have been developed over the years, mainly adding dynamic obstacles in the grid.
	\citet{Saval-Calvo2017} wrote a review of the different Bayesian Occupancy Filter frameworks, presenting a taxonomy of the methods.
	\citet{OCallaghan2012} proposed a way to store the occupancy map without discretization by using Gaussian processes.
	This method keeps the dependence between cells in the grid, which was not the case in the original occupancy grid from \citet{Elfes1989}. 
	In an attempt to reduce the complexity of the Gaussian process,~\citet{Kim2013} used overlapping local Gaussian processes. 
	Regardless of this amelioration, the~time computational complexity is still an issue for these kinds of methods, whereas the standard occupancy grids, as well as our method, do not suffer from such problems.
	\citet{Ramos2016} developed an analog method using Hilbert maps (HMs), overcoming the computational complexity of the Gaussian process.
	In order to take into account the uncertainty in the different parameters, the~method was  extended to Bayesian Hilbert maps by \citet{Senanayake2017}.
	Lately, HMs have been generalized to dynamic environments by \citet{Guizilini2019}, allowing real-time occupancy predictions. 
	These methods still need to tune parameters that have great consequences on the quality of the resulting maps.
	An alternative algorithm for occupancy grids was  presented in \citet{Agha-mohammadi2019} by storing richer data in a map, taking into account the estimation of the variance for each cell.
	Under these considerations, our framework and the Bayesian occupancy grid are very alike, as~the environment is tessellated into cells.
	The Lambda Field also stores a confidence interval over each cell in the same fashion as \citet{Agha-mohammadi2019}.}

	Once a representation of the environment is available, the~robot can start to assess risk in its map.
	The risk was first defined by the likelihood of not colliding with anything, as suggested by \citet{Fraichard2007}.
	In the context of autonomous driving, the~Time to Collision introduced by \citet{Lee1976} is widely used.
	This metric measures the time at which the robot will collide with a specific obstacle given the current path.
	The Time to Collision is useful in accident mitigation systems, but is not well fitted for long-term planning.
	It is mainly useful for mitigating the speed of a vehicle in traffic.
	It has been demonstrated by \citet{Laugier2011} that the Time to Collision lacks context, and is hence  not the best solution for every situation.
	Furthermore, this kind of metric is used in the context of known dynamic obstacles and is not easily transposable for path-planning algorithms.
	Given these factors, the~risk has to be defined in another fashion in the context of path planning in occupancy~grids.
	
	The risk is also very dependent on the application.
	For example, \citet{Vaillant1997} and \citet{Caborni2012} tackled the problem of path planning for neurosurgery.
	For this application, the~risk depends essentially on which zones of the brain the tool goes through.
	It is very different from standard path-planning risks, as in this context, we are sure to collide with a part of the brain for every possible path.
	In that sense, the~path-planning framework should be able to tackle different types of risk, which is the case of the~Lambda Field.
	
	\citet{Majumdar2017} addressed the issue of how a robot should quantify risk and what constitutes a `good' risk metric.
	They came to the conclusion that a risk measure is said to be coherent if it satisfies axioms, showing that otherwise, the~risk metric can have undesired behaviors.
	However, the~physical meaning is neglected, which leads to non-intuitive definitions for risk metrics and difficult parameter settings for path planning.
	In our work, we thereupon define the risk as an understandable quantity, which is, for our application, the expected force of collision on a given path.
	Our metric also respects the axioms defined by \citet{Majumdar2017}, leading to a risk proven to behave as desired, as well as having a physical unit.
	\textls[-10]{In the context of occupancy grids, a~risk map is widely used to deal with the risk.
	For path planning, the~occupancy grid is replaced with a risk map, where each cell stores the risk at this position.
	The higher the risk, the~more the robot should avoid this place while planning its trajectory.
	\citet{Tsiotras2007} used wavelets to store the environment and a risk map at different scales;
	for this application, the~risk was defined as the probability of occupancy.
	Then, the~path-planning method was to find the path minimizing the overall sum of the risk of each traversed cell.
	Hence, the~total risk lacks physical meaning.
	Our framework differs from the previous one, as the total risk of a path has the same unit as the risk metric used.
	In another context, \citet{DeFilippis2011} used a risk map to control the altitude of an unmanned aerial vehicle.
	The risk was defined as the probability of flying in unsafe conditions for a given point on the map.
	In the same fashion,~\citet{Primatesta2019} defined the risk as the hourly probability of lethal incidents for each position of the unmanned aerial vehicle.
	The risk was also set to the maximum for obstacles and no-fly zones.
	\citet{Joachim2008} used a risk map to prevent a robot from going too close to dangerous obstacles, such as pedestrians or other cars, allowing the robot to safely navigate in narrow spaces, such as parking \mbox{spots.~\citet{Pereira2011}} also used a risk map to find the best path for underwater vehicles, where the risk was set to the probability that the position was occupied by an obstacle.
	Although all these methods demonstrated good results, they all assumed that the risk was only a function of the position, omitting the robot configuration.
	As said before, the~robot configuration can greatly change the risk.
	For instance, going to a position at high speed is often more dangerous than going at a low speed.
	\citet{Feyzabadi2014} defined a risk function that depends on the position as well as the robot action.
	As a result, the~robot could choose to go to a position only if its speed was low enough.
	Our framework uses the same idea while giving a physical meaning to the cost of the overall path.
	Therefore, the~probability of collision is a metric that is too simple to perfectly describe the risk.
	As said by \citet{Eggert2014}, in~the case of ADAS systems, we would rather want to assess the expected damage done to the vehicle than the probability of collision.
	We then propose a framework allowing the computation of a generic risk that can be defined depending on the application.}
	Using its representation of the environment and a risk function, the~robot can start planning.
	Many of the popular methods use a binary representation of the environment, meaning that any point in the environment is either free or occupied.
	A review of such algorithms can be found in \citet{Tsardoulias2016}. %
	The most common way to convert the Bayesian grid into a binary grid is to apply a user-defined threshold, as done \mbox{by \citet{Yang2013}.}
	However, applying a threshold to the environment might lead to discarding some obstacles, commanding the robot to plan entry into potentially occupied zones.
	A review of algorithms of path planning in occupancy grids was  done \mbox{by \citet{Cikes2011}.}
	The different algorithms presented in this article all aim to minimize the cost function of the path, which is the sum of the cost of each traversed cell. 
	The cost of a path has no physical meaning; thus, determining if the path is truly safe might become a difficult~task.

	Another method proposed by \citet{Fulgenzi2007} is to cluster the occupancy grid, leaving the unclustered space as free or occluded.
	The risk assessment is then reduced to evaluating the risk for probabilistic known obstacles.
	However, such clustering can be very difficult to compute in unstructured environments.
	We thus need a way to evaluate the cost of a path in occupancy grids while taking into account the probability of occupancy.
	Using Rapidly Exploring Random Trees, \citet{Fulgenzi2008} and \citet{Fulgenzi2009} defined the cost of a path as the joint probability of not having a collision in each node.
	Their framework assumes that traveling between nodes is risk-free; 
	if we do not make this assumption, we fall back on the initial problem of computing a cost over a continuous path.
	To overcome the problem of computation of the risk over a continuous path, several methods have been proposed.
	\citet{Rummelhard2014} defined the risk in a Bayesian occupancy grid as the maximum probability of collision over the cells. 
	Nevertheless, there is no natural way to include a more complicated risk in the framework, and these metrics can show unintended behaviors in complicated scenarios.
	Indeed, traversing one high-probability cell has the same risk as traversing ten cells of the same probability for the second \mbox{metric. 
	\citet{Dhawale2018}} chose to represent the environment as a Gaussian process and represented the obstacles using a threshold over the Gaussians.
	Doing this dissociates free space and occupied space, falling back on the methods presented \mbox{in \citet{Tsardoulias2016}.}
	\citet{Gerkey2008} computed the cost of a path by summing the probability of occupancy of the cells that the path crosses.
	This sum is then injected into a global cost function, taking into account other constraints, such as the speed or the distance to the objective, where each constraint has a user-defined coefficient.
	\citet{Francis2018} used the same idea for path planning in Hilbert maps, which were introduced by \citet{Ramos2016}.
	The drawback of these methods is that the cost lacks physical meaning, as they sum probabilities.
	Since this sum does not have any physical unit, its associated coefficient does not have one either, making its tuning non-intuitive for the user.
	Finally, \citet{Heiden2017} used the concept of the product integral to compute the probability of collision over a path.
	It leads to a probability of collision, but~this method has no physical meaning.
	We show in this article that our framework can be seen as the generalization of their framework.

\section{Theoretical~Framework} \label{sec:theory}
				We present in this section the theoretical framework for assessing a generic risk over a path in Lambda Fields.
				First, we justify the use of the mathematical tools by showing how they naturally arise while dealing with continuous environments.
				Then, we address the construction of the Lambda Fields in Section \ref{subsec:field}, as~well as a way to compute confidence intervals over the field in Section \ref{subsec:CI}. 
				Indeed, the~more the cells are measured, the~more confident the robot should be to move. 
				Next, we present in Section \ref{subsec:framework} a framework capable of assessing a generic risk over a path in a Lambda Field.
				We then extend this framework and design a risk function allowing the robot to navigate in unstructured environments, such as tall grass, in Section \ref{subsec:masses}.
				Finally, we improve the framework of \citet{Heiden2017} in \mbox{Section \ref{subsec:reachability}} to take into account the size of the robot and show that, under~our improvement, it can be seen as a special case of our framework.

				The key concept of the Lambda Field is its ability to assess the probability of collision inside a subset of the environment (e.g., the path of the robot), leading to the computation of a generic risk that can be adjusted depending on the scenario. %
				To better understand the reasons for the following framework, we will first demonstrate its construction.
				We assume that the probability of encountering a collision for a path of area $\Delta a$ is $\lambda_i \Delta a$, where $\lambda_i\in\mathbb{R}_{\geq 0}$ is the rate of the event `collision' and $\Delta a \rightarrow 0$ such that $\lambda_i\Delta a \leq 1$.
				The larger the intensity $\lambda_i$ is, the~more likely it is that a collision will occur.
				In a macroscopic approach, the~intensity $\lambda_i$ corresponds to the expected number of collisions in a cell of area $\SI{1}{\m\squared}$ and can, therefore, vary from $0$ (i.e., the cell will never create a collision) to $+\infty$ (i.e., the cell will create an infinite number of collisions during the traversal).

				The probability of crossing $N$ surfaces of areas $\Delta a$ with a rate $\lambda_i$ without collision is
\begin{equation}
					\prod_{i=0}^{N-1} (1-\lambda_i\area).
				\end{equation}
				
				Taking the limit of the path area $\Delta a \rightarrow 0$ leads to the computation of the Volterra type I product integral.
				For a path crossing a total area of $A$ where each subregion of area $\Delta a$ has a rate $\lambda(a)$, $a$ being the total area crossed from the beginning, we have
\begin{equation}
					\lim_{\Delta a\rightarrow 0} \prod_{i=0}^{A/\area} (1-\lambda(i\area)\area) = \expp{-\int_0^A \lambda(a) \dif a}.
				\label{eq:prod_int}
				\end{equation}
				
				A proof of Equation (\ref{eq:prod_int}) can be found in~\cite{Slavik2007}.
				The probability of encountering no collisions over a path is then the probability that no event `collision' happens in a heterogeneous Poisson point distribution of rate $\lambda(a)$.
				Taking the limit of a binomial distribution indeed leads to a Poisson point process distribution.
			Hence, the~natural way of dealing with collisions in a continuous manner is to use a Poisson point process distribution.
			This process counts the number of events that have happened given a certain area, depending on the mathematical space.
			In our case, we want to count the number of the `collision' events that could occur given a path (i.e., a subset of $\mathbb{R}^2$). 
			We point out that the theory is here presented for 2D paths, but~the extension in $\mathbb{R}^3$ is trivial, as~the only change is the tessellation of the map being in 3D instead of 2D.
			For a positive scalar field $\lambda(\bm{x})$, with~$\bm{x} \in \mathbb{R}^2$, the~probability of encountering at least one collision in a path $\mathcal{P}~\subset~\mathbb{R}^2$ is
\begin{equation}
				\pr{ \mathtt{coll}|\mathcal{P} } = 1 - \expp{-\int_\mathcal{P}\lambda(\bm{x})\dif \bm{x}}.
				\label{eq:collision_proba}
			\end{equation}

			{Nonetheless, it is impossible to both compute and store the field $\lambda(\bm{x})$, as it has an infinite number of degrees of freedom.
                        Hence, we tessellate our field into cells of a fixed size in a fashion similar to that of Bayesian occupancy grids.
                      Throughout this paper, we will assume that we are dealing with a tessellated field where each cell has an area $\area\in\mathbb{R}_{>0}$.}
			By tessellating the field, the~probability of collision is given by
\begin{equation}
                          \pr{ \mathtt{coll}|\mathcal{P} } = 1 - \expp{-\Lambda(\mathcal{C})} \quad\text{with } \Lambda(\mathcal{C}) = \area\sum_{c_i\in\mathcal{C}}\lambda_i , 
				\label{eq:approxColl}
			\end{equation}
                        for a path $\mathcal{P}$ crossing the cells $\mathcal{C}=\{c_i\}_{0:N-1}=\{c_0,\dots,c_{N-1}\}$, where each cell $c_i$ has an area of $\area$ and an associated lambda $\lambda_i$, which is the intensity of the cell. 
			The lambda can be seen as a measure of the density of the cell: The higher the lambda is, the~more likely it is that a collision will happen in this~cell.

			Using this representation, we hereby see that the probability of collision is not dependent on the size of the cells.
			It is indeed the same to compute the probability of collision for crossing two cells of area $\area/2$ or one cell of area $\area$ for a constant $\lambda$.

	\subsection{Computation of the~Field}
		\label{subsec:field}
\begin{figure}[H]
					
					\includegraphics[width=.9\linewidth]{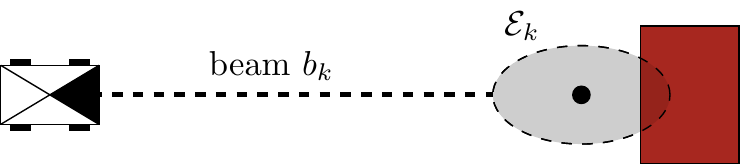}
					\caption{The robot measures an obstacle using a \ac{lidar} sensor. The~obstacle (in red) is in the area $\mathcal{E}_k$ (in gray) centered on the measurement (black dot).}
					\label{fig:errorRegion}
			\end{figure}

		As we established a new approach to representing the occupancy of an environment, we need to develop a way to dynamically compute the lambdas. 
		We assume that the robot is equipped with a \ac{lidar} sensor, which gives a list of cells crossed by beams without collision and~another list of cells where the beams collided.
		Using this sensor model, we construct the Lambda Field in the following manner.
		We want to find the combination of $\lambda=\{\lambda_i\}_{0:M_C-1}$ for~a map tessellated into $M_C$ cells that maximizes the expectation of the $K$ beams that the \ac{lidar} has shot since the beginning. 
		In addition, each \ac{lidar} beam has an associated error region $\mathcal{E}_k$ of area $e_k$ centered on the measurement, meaning that the actual obstacle is in $\mathcal{E}_k$.
		Figure \ref{fig:errorRegion} shows an example of such a \ac{lidar} beam error~region.

		Therefore, each lidar collision gives a region where an obstacle is.
		This kind of sensor simplification is common and was used, for example, by~\cite{Rohou2018}.
		At this stage, we assume that every lidar measurement possesses the same error region area $e$.
		The case where each beam has a different error region is covered in Appendix \ref{appendix:heterogeneousErrRegions}, which can be useful for radar measurements or lidars with substantial beam divergence.
		For each lidar beam $b_k$, the~beam crossed the cells $c_m\in \mathcal{M}_k$ without collision and hit an obstacle contained in the cells $c_h\in\mathcal{E}_k$.
		The log-likelihood of the beam $b_k$ is
\begin{equation}
		\mathcal{L}(b_k|\lambda) = \ln\left[ \expp{-\Lambda(\mathcal{M}_k)}\left( 1-\expp{-\Lambda(\mathcal{E}_k)}\right) \right].
		\end{equation} 

		The log-likelihood of $K$ lidar beams is then
\begin{equation}
					\begin{aligned}
						&\mathcal{L}(\{b_k\}_{0:K-1}|\lambda) = \sum_{k=0}^{K-1} \mathcal{L}(b_k|\lambda) \\
												&= \sum_{k=0}^{K-1} \left[ -\Lambda(\mathcal{M}_k) + \ln\left(1- \expp{-\Lambda(\mathcal{E}_k)}\right)\right].
					\end{aligned}
					\label{eq:logLikelihood}
			\end{equation}
			
			We want to maximize this quantity and, hence, nullify its derivative, as the function is concave.
			In order to find a closed form, we approximate the derivative with the assumption that the variation of lambda inside the error region of the lidar is small enough to be negligible.
			Thus, for~each $\lambda_i\in\mathcal{E}_k$, we have
\begin{equation}
				\area\sum_{c_h\in\mathcal{E}_k}\lambda_h \approx e\lambda_i.
				\label{eq:mapping_approx}
			\end{equation}
			
			Using this approximation, the~derivative is
\begin{equation}
					\frac{\partial \mathcal{L}(\{b_k\}_{0:K-1}|\lambda)}{\partial \lambda_i} \approx -m_i\cdot\area + h_i\frac{\area}{\exp(e\lambda_i)-1},
			\end{equation}
			where $m_i$ is the number of times that the cell $c_i$ has been counted as `miss' (i.e., was outside the error region), and $h_i$ is the number of times the cell $c_i$ has been counted as `hit' (i.e., was in the error region of the sensor).
			We finally find the zero of the derivative, leading to
\begin{equation}
					\lambda_i = \frac{1}{e}\ln\left(1+\frac{h_i}{m_i}\right).
					\label{eq:lambda}
			\end{equation}

			This closed form allows a low computational complexity of the Lambda Field.
			We also see that the formula is independent of the size of the cells, which is the main limitation of the current representation that we were aiming at~resolving.

			We are then able to construct the Lambda Field using Equation (\ref{eq:lambda}).%

			\subsection{Confidence~Intervals}
				\label{subsec:CI}
			In the same way as~\cite{Agha-mohammadi2019}, we define the notion of confidence over the values in the Lambda Field.
			Indeed, the~more the cells are measured, the~greater the confidence in the robot's movements should be.
			For each cell $c_i$, we seek the bounds $\lambda_L$ and $\lambda_U$ such that
			
\begin{equation}
				\begin{aligned}
					&\mathbb{P}(\lambda_L \le \lambda_i \le \lambda_U) \ge \SI{95}{\percent}\\
					\Leftrightarrow &\mathbb{P}(\lambda_L \le \frac{1}{e}\ln\left(1+\frac{h_i}{m_i}\right) \le \lambda_U) \ge \SI{95}{\percent}.
				\end{aligned}
			\end{equation}
			
			To compute those bounds, we introduce the notion of false positives and false negatives:
			Every cell measurement $j$ has a probability $p_j^h$ of rightfully reading `hit' and a probability $p_j^m$ of rightfully reading `miss'.
			The probabilities $p_j^h$ and $p_j^m$ have to be experimentally computed and can vary according to a great number of parameters; for example, the~probability $p_j^h$ is lower in the event of heavy rain or snow.

			Using the relation $h_i = M - m_i$, where $M$ is the number of times the cell has been measured, we can rewrite the above equation as
\begin{equation}
				\mathbb{P}(K_L \le h_i \le K_U) \ge \SI{95}{\percent},
			\end{equation}
			such that
\begin{equation}
				\begin{aligned}
					\lambda_L &= \frac{1}{e}\ln\left(\frac{K_L}{M-K_L}+1\right), \\
					\lambda_U &= \frac{1}{e}\ln\left(\frac{K_U}{M-K_U}+1\right).
				\end{aligned}
				\label{eq:KLKU}
			\end{equation}

			The quantity $h_i$ can be seen as a sum of $M$ Bernoulli distributions, such that
\begin{equation}
				\begin{aligned}
					h_i &= \sum_{j=0}^{h_i-1} \bar{h}_j + \sum_{j=0}^{m_i-1} \left(1-\bar{m}_j\right), %
				\end{aligned}
			\end{equation}
			where $\bar{h}_j$ and $\bar{m}_j$ are Bernoulli variables equal to 1 if the reading was right and 0 otherwise. 
			The quantity $\sum_j\left(1-\bar{m}_j\right)$ is hence the number of times the sensor wrongfully reads `hit' instead of `miss'.

			The distribution of $h_i$ is not binomial, but~a Poisson binomial distribution with poor behaviors in terms of computation. 
			Since the Poisson binomial distribution satisfies the Lyapunov central limit theorem, we can approximate its distribution with a Gaussian distribution of same mean and variance:
\begin{equation}
				\begin{aligned}
					\mu &= \sum_{j=0}^{h_i-1} p_j^h + \sum_{j=0}^{m_i-1} 1-p_j^m \quad \text{and} \\
					\sigma^2 &= \sum_{j=0}^{h_i-1} p_j^h(1-p_j^h) + \sum_{j=0}^{m_i-1} p_j^m(1-p_j^m).
				\end{aligned}
			\end{equation}
			
			We can then have the bounds at 95\%, for example, with~\begin{equation}
					\begin{aligned}
						K_L &\approx \max(\mu-1.96\sigma, 0) ,\\
						K_U &\approx \min(\mu+1.96\sigma, M). %
					\end{aligned}
			\end{equation}
			
			\textls[-15]{The bounds $\lambda_L$ and $\lambda_U$ are then retrieved from $K_L$ and $K_U$ using Equation (\ref{eq:KLKU}).}

		\subsection{Generic Framework for Risk~Assessment}
			\label{subsec:framework}
			As mentioned before, the~motivation for the Lambda Field is its ability to compute path integrals and, hence, a risk along a path.
			This risk can be defined depending on the application and is independent of the following framework, meaning that it can be interchanged without any modification of the theory.
			For a path $\mathcal{P}\subset \mathbb{R}^2$ crossing the cells $\mathcal{C}=\{c_i\}_{0:N-1}$ in order, the~probability density function (p.d.f) over the Lambda Field is
\begin{equation}
          f(a) = \expp{n\Lambda\left(\{c_n\}\right)-\Lambda\left(\{c_j\}_{0:n-1}\right)} \cdot \lambda_n\expp{-a\lambda_n},
			\label{eq:pdflf}
	\end{equation}
	where $n=\floor{a/\area}$, $\floor{\cdot}$ is the standard floor function and $\Lambda(\cdot)$ is defined in Equation (\ref{eq:approxColl}).
	The variable $a$ denotes the area the robot has crossed. 
        Note that $\{c_n\}$ is a singleton, i.e., $\Lambda(\{c_n\})=\area\lambda_n$, whereas $\Lambda(\{c_j\}_{0:n-1})=\area\sum_{i=0}^{n-1}\lambda_i$ sums $n$ elements (and equals zero if $n=0$).

	One can note that the conversion of the area $a$ into the curvilinear abscissa is trivial, the latter being more convenient for path-planning applications.
	For a robot of width $W$ that has crossed an area $a$, its curvilinear abscissa $s$ equals  $a/W$.
        {The length of the robot is not considered, as we assume that the body of the robot only spans cells that have already been crossed.
        As such, only the front of the robot of width $W$ discovers new cells that are potentially risky.}

	Furthermore, Equation (\ref{eq:pdflf}) can be easily proved, as integrating $f(a)$ over a path $\mathcal{P}$ crossing the cells $\mathcal{C}=\{c_i\}_{0:N-1}$ in order gives the probability of encountering at least one~collision:
\begin{equation}
		\pr{\mathtt{coll}|\mathcal{P}} = \int_0^{N\area} f(a)\dif a = 1 - \expp{-\Lambda(\mathcal{C})}.
	\end{equation}
	
	We can then define the expectation of a risk function $r(\cdot)$ over the path as
\begin{equation}
		\mathbb{E}[r(A)] = \int_0^{N\area} f(a) r(a) \dif a .
	\end{equation}
	
	The random variable $A$ denotes the crossed area at which the first `collision' event occurs.
	If the cells are small, we can assume that the function $r(\cdot)$ is constant inside each cell.
	Using this assumption, we simplify the above equation to
\begin{equation}
		\mathbb{E}[r(A)] = \sum_{i=0}^{N-1} K_i r(\area i) \quad \text{with } K_i = \expp{-\Lambda(\{c_j\}_{0:i-1})}\left[1-\expp{-\Lambda(\{c_i\})}\right],
		\label{eq:risk}
	\end{equation}
	for a path $\mathcal{P}$ going through the cells $\{c_i\}_{0:N-1}$.

	The risk function $r(\cdot)$ is generic and can take into account the state of the robot, as well as the state of the world.
	One can notice that the special case $r(\cdot)=1$ leads to the probability of collision given by Equation (\ref{eq:approxColl}).
	Furthermore, the~probability density $f(a)$ only looks at the risk generated by the first collision occurring on the path.
	Therefore, it is assumed that the robot stops after any collision and does not continue its course.
	This assumption can be lifted if necessary, as~shown in the next~section.

	For our applications, we chose to model the risk as the force of collision (i.e., loss of momentum) if the collision occurs at the area $a$.
	It is indeed a good quantification of the damage induced by the collision and is a better metric of the risk than the probability of collision, as~shown by \citet{Eggert2014}.
	First, we present as an example a way to assess this risk by assuming that every obstacle has an infinite mass.
	Indeed, this assumption holds for most scenarios where the robot's mass is negligible compared to the obstacles' masses (e.g., a tree or a wall).
	We then lift this assumption in Section \ref{subsec:masses}, where each obstacle now has a probabilistic mass, allowing the robot to evolve in unstructured~environments.

\begin{figure}[H]
		
		\includegraphics[width=.5\linewidth]{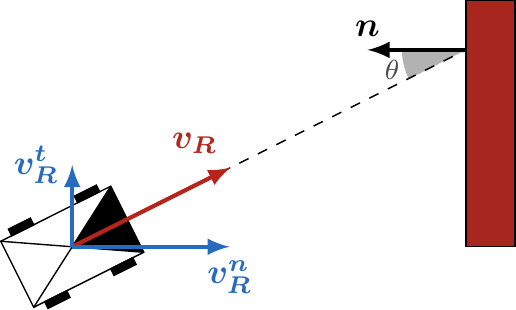} 
		\caption{
                  A robot
                        of speed $v_R = ||\bm{v_R}||$ collides with an obstacle of normal $\bm{n}$ with an angle $\theta$. The~speed of the robot can be decomposed into the tangential component $\bm{v_R^t}$ and the normal component $\bm{v_R^n}$. Only the latter influences the collision with the obstacle.
		}
		\label{fig:CollNormal}
	\end{figure}

	Assuming the obstacle that the robot collides with has an infinite mass, the~force of collision is computed as
\begin{equation}
		r(a) = m_R\cdot v_R^n(a),
		\label{eq:qttmvt}
	\end{equation}
	where $m_R$ is the mass of the robot, and~$v_R^n(a)$ is its velocity towards the obstacle at the area $a$. 
	As shown in Figure \ref{fig:CollNormal}, the~velocity towards the obstacle of normalized normal $\bm{n}$ is
\begin{equation}
	\begin{aligned}
		v_R^n &= \left| \bm{n}^T\bm{v_R} \right| \\ 
			  &= \left|v_R \cdot \cos(\theta)\right| \qquad\text{ for } ||\bm{n}||=1
	\end{aligned}
	\end{equation}
	where $\cdot^T$ stands for the usual vector transpose, $v_R = ||\bm{v_R}||$ the robot velocity, and $\theta$ the angle between the robot heading and the obstacle's normal.
	The angle of collision is interesting to take into account for numerous scenarios, such as an autonomous vehicle driving over a cliff.
	Because of skidding, the~vehicle may find itself in a configuration where it has no choice but to collide with the safety railing. 
	The best choice will intuitively be to minimize the collision and, hence, collide with the railing with a high incidence angle.%

	This risk metric assumes that every obstacle the robot might encounter has an infinite mass. 
	We also assume perfect inelastic collision, as~most deployed vehicles are designed to absorb collisions as much as possible.
	This means that if the robot collides with an obstacle, the~resulting collision would lead the robot to stop (i.e., losing a momentum of $m_R\cdot v_R^n$). 
	Depending on the application, other metrics can be developed.
	We present in the next section the development of a more complicated metric allowing the robot to navigate through unstructured obstacles such as tall grass by lifting the approximation that all obstacles have infinite~masses.

\subsection{Taking into Account the Mass of the~Obstacles}
\label{subsec:masses}
	In the context of autonomous navigation, the~robot might have to go through objects that look like obstacles from the point of view of the lidar, but~are in fact harmless for the robot.
	An ideal example of this scenario is where the robot has to go through tall grass to reach its goal.
	Since the lidar returns very close measurements of the grass around the robot, the~robot would be unable to move.
	However, with images provided by a camera, an~algorithm could clearly detect that the obstacles are only tall grass; hence, the robot should proceed and reach its goal.

	As the risk metric developed in the previous section assumes that every obstacle has an infinite mass, it is unable to deal with such scenarios.
	Thus, this assumption is lifted, and each obstacle is assumed to have a probabilistic mass.
	We thereby estimate the class of the obstacles in each cell and infer the associated probabilistic mass distribution.
	This can be done with a camera and deep learning segmentation~\cite{Badrinarayanan2017} or radar classification~\cite{Lalonde2006}.
	In addition to the Lambda Field, we store a map of the probability distribution function of the mass distribution for each cell, which is provided by one of the above-cited methods. 
	Furthermore, as~collisions with low-mass obstacles do not pose a threat to the robot, the~risk metric is defined as the force of collision with obstacles that will stop the robot, therefore discarding threat-less~collisions.

\begin{figure}[H]
		
		\includegraphics[width=.9\linewidth]{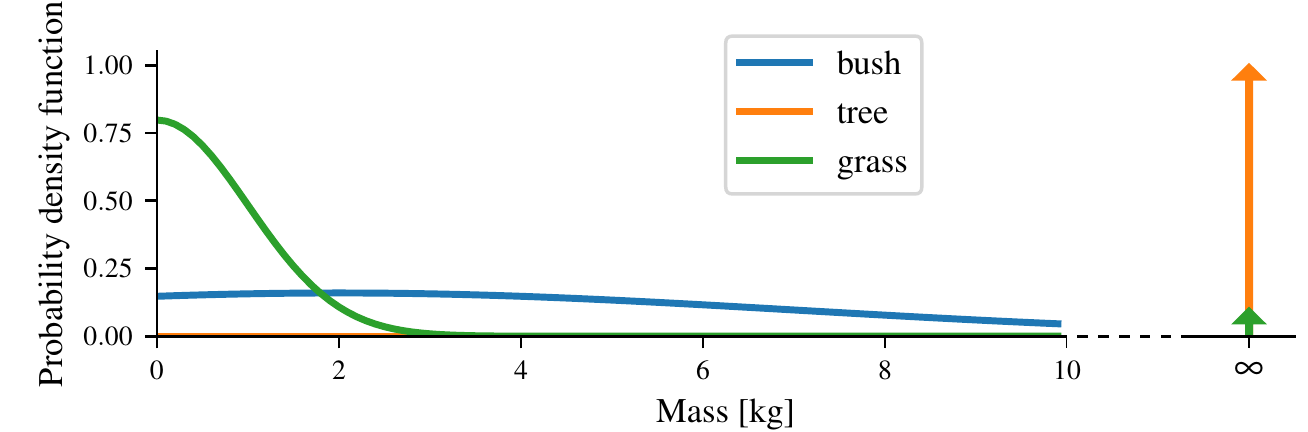}
		\caption{Examples of probability density functions of several labels. 
				 The arrow represents the Dirac delta function.
				 The mass of the grass is very likely to be close to zero, but there is a chance that a high-mass obstacle is hiding in it (e.g., a rock).
				 The mass of a bush is very uncertain, as it may be more or less dense.
				 In contrast, the~mass of a tree is always very high.
				}
		\label{fig:masses_pdf}
	\end{figure}

	Figure \ref{fig:masses_pdf} shows examples of probability distribution functions for several obstacle classes.
	The main use of a probabilistic formulation for the masses is to deal with the uncertainty of the labels.
	Indeed, the~grass can easily hide a high-density obstacle, such as a rock.
	Moreover, the~mass of the vegetation is very variable, and the robot can expect a harmless collision as much as a harmful collision while going through these kinds of obstacles. 
	In the case where no label is available for a cell, the~worst case is taken into account, meaning that the mass of the cell is set to~infinity.

	We chose to discretize the probability density function into a sum of Dirac impulsions $\delta(\cdot)$.
	The mass p.d.f $f^m_i(\cdot)$ of the cell $c_i$ is then
\begin{equation}
		f^m_i(m) = \sum_{k=0}^\infty \alpha_{ik} \cdot \delta(m - k\Delta m) ,
	\end{equation}
	with $\Delta m$ being the discretization step and $\alpha_{ik}$ being the probability that $m\in[k\Delta m, (k+1)\Delta m]$.
	In addition, only a finite number of $\alpha_{ik}$ are not null in order to store the~p.d.f.

	A problem quickly arises from Equation (\ref{eq:risk}) if we want to take into account the mass of the obstacles.
	The equation only looks at the first collision, as it assumes that any collision would lead the robot to stop its course.
	For very light obstacles, such as grass, this assumption falls apart.
	Hence, we need to add a term to the equation to allow the robot to continue its course after a collision.
	To do so, we need to understand the meaning of the lambdas.
	For an area $\area$ where the Lambda Field is constant with a value $\lambda$, the~expected number of `collision' events is $\area\lambda$.
	Using the probability $p_{si}$, the~probability of the robot being stopped because of the collision at the cell $c_i$, we want that each collision has the probability $p_{si}$ of being harmful for the robot.
	Hence, we use our probability of traversal $p_{si}$ as a new measure over the field. %
	Given the harmful probability $p_{si}$ for the cell $c_i$, the~intensity function $\Lambda(\mathcal{C})$ becomes
\begin{equation}
		\Lambda_m(\mathcal{C}) = \area\sum_{c_i\in\mathcal{C}}\lambda_ip_{si}.
	\end{equation}
	
	Using this newly defined intensity measure, only hazardous collisions are investigated, treating collisions that do not stop the robot as~harmless.

	Assuming that the robot can go through obstacles if their mass is below a certain threshold $m_\mathtt{max}$, the~probability $p_{si}$ is then
\begin{equation}
	\begin{aligned}
		p_{si} &= \mathbb{P}\left(m_i > m_\mathtt{max}\right) \\
			   &= 1 - \int_0^{m_\mathtt{max}} f^m_i(m)\dif m ,
	\end{aligned}
	\end{equation}
	where $f^m_i(\cdot)$ is the p.d.f of the mass of the cell $c_i$.
	In addition, since we do not assume anymore that the obstacles have infinite masses, the~risk $r(\cdot)$ (i.e., the loss of momentum at the impact) becomes
\begin{equation}
	\begin{aligned}
		r_m(\area i, m) &= m_R\left(v_R^n(\area i)-\frac{m_Rv_R^n(\area i)}{m_R+m}\right)  \\
				   &= m_R \frac{m\cdot v_R^n(\area i)}{m_R+m},
	\end{aligned}
	\end{equation}
	where $m$ is the mass of the $i$th cell.
	Since the mass of each obstacle is probabilistic, we need to sum over all the possible masses to find the expected force of collision over a path, leading to (the proof is detailed in Appendix \ref{appendix:proofExpCollMass}):
\begin{equation}
	\begin{aligned}
		\mathbb{E}[r_m(A, M)] &= \sum_{i=0}^{N-1} K_i \int_0^\infty f^m_i(m)r_m(\area i,m) \dif m \\
						 &= \sum_{i=0}^{N-1} K_i \sum_{k=0}^\infty \alpha_{ik} r_m(\area i, k\Delta m)  .\\
	\end{aligned}
		\label{eq:expected_coll_masses}
	\end{equation}
	where $M$ is the random variable corresponding to the mass of the cell in which the collision happened, and~$K_i$ is computed in the same way as in Equation (\ref{eq:risk}), but using $\Lambda_m(\cdot)$.
	One can note that we can rewrite the above equation in the form
\begin{equation}
	\begin{aligned}
		\mathbb{E}[r_m(A, M)] &= \sum_{i=0}^{N-1} K_i\, r'(\area i) \\
		\text{with }\quad r'(\area i) &= \sum_{k=0}^\infty \alpha_{ik} r_m(\area i, k\Delta m),
	\end{aligned}
	\end{equation}
	hence going back to the known expectation formula of  Equation (\ref{eq:risk}).
	We omitted the parameters $\alpha_{ik}$ in the parameters of $r'(\cdot)$, as they are directly retrievable from the crossed area $\area i$.
	In addition, notice that setting ${\mathbb{P}(m_i=\infty)=1}$ for all of the cells leads, as expected, to the same risk as when using Equation (\ref{eq:qttmvt}).

        {One can note that there is no direct way of taking into account the mass of the obstacles in the Bayesian occupancy grids.
        Indeed, the~Bayesian occupancy grid only stores the information of occupancy for a given cell instead of  more abstract information, that is, the intensity of the `collision' event in the case of the Lambda Field.
        As such, the~Lambda Field possesses an extra layer of assessment where, using the risk function, the~framework quantifies the risk associated with the event.
        This layer allows one to take into account numerous  types of information, such as the mass of the obstacles, as~done in this section.
      Furthermore, the~Bayesian occupancy grid suffers from its dependence on the tessellation size and is, thus, not suited to inferring generic risks on a path, as~stated in the introduction.}

      In the following, we analyze a method for assessing the probability of collision in Bayesian occupancy grids that does not depend on the tessellation size~\cite{Heiden2017}.
        We show that, with our improvement to take into account the size of the robot, their method can be seen as a special case of our~framework.

		\subsection{Comparison and Improvement of the Reachability~Metric}
		\label{subsec:reachability}
			In this section, we analyze and adapt the concept of \emph{reachability} defined in~\cite{Heiden2017}.
			These authors' work was indeed the first to address the problem of risk assessment in occupancy grids, which is shown in Figure \ref{fig:intro}.
			We first investigate the different metrics proposed in the article and then show that, with our improvement to consider the size of the robot, our framework can be seen as a generalization of their method.
			They propose the use of the concept of the product integral, which is the product counterpart of the standard integration.
			A summary of the product integration can be found in \citep{Slavik2007}.
			They introduced the probability of occupancy $p_o(\cdot)$ (defined in their article as $m(\cdot)$) as the density of the cell.
			At first, they defined the reachability $R_t$ for a path from the time $t=0$ to $T$ as a product integral, computed as
\begin{equation}
				R_t = \prod_0^T (1-p_o(x(t)))^{\dif t},
			\end{equation}
			where $x(t)$ is the robot position at the time $t$ and $p_o(x(t))$ is the probability that the position $x(t)$ is occupied.
			The higher the reachability, the~safer the corresponding path is.
			However, they argue that it would be better to consider the distance traveled through a cell instead of the time.
			It is indeed better, as the first metric leads to a counter-intuitive reachability:
			for a robot crossing at a speed $v$ a straight path of length $l$, where all cells have the probability $p_o$ of being occupied, the~reachability is
\begin{equation}
			\begin{aligned}
				R_t	&= \prod_0^{l/v} (1-p_o)^{\dif t} \\
					&= \lim_{\Delta t \rightarrow 0} \prod_{i=0}^{l/v/\Delta t} (1-p_o)^{\Delta t}. \\
			\end{aligned}
			\end{equation}
			
			Using the fact that $(1-p_o)^{\Delta t} = \expp{\ln(1-p_o)\Delta t}$ and the Riemann definition of the integral, the~expression can be simplified to
\begin{equation}
			\begin{aligned}
				R_t	&= \expp{\int_0^{l/v}\ln(1-p_o)\dif t} \\
					&= (1-p_o)^{l/v}.
			\end{aligned}
                        \label{eq:Rt}
			\end{equation}
			
			{The reachability from the first metric $R_t$ is then higher when the speed is high, meaning that it would be safer to cross the path at a higher speed.
                        Indeed, the~trajectory is parametrized on the time of traversal; as such, the~faster the robot is going, the~smaller the number of position samples to evaluate will be.
                      A robot of infinite speed would thus consider all paths safe, as the integration would not carry any sample points, whereas a very slow robot would lead to consideration of more sample points and, therefore, lower the reachability, as~shown in} Equation (\ref{eq:Rt}).

			Their second reachability metric $R_L$ does not possess such a behavior, as they parametrized the integral over the traveled distance $L(t, t+\dif t)$ between two instants, leading to
\begin{equation}
			\begin{aligned}
					R_L &= \prod_0^T (1-p_o(x(t)))^{L(t, t+\dif t)} \\
					    &= \prod_0^T (1-p_o(x(t)))^{|\dot{x}(t)|\dif t}.
			\end{aligned}
			\label{eq:R_L}
			\end{equation}
			
			Since the traveled distance $d(t)$ equals  $\int_0^t |\dot{x}(t)|\dif t$, we have $\dif d(t) = |\dot{x}(t)| \dif t$, and Equation (\ref{eq:R_L}) can be simplified to
\begin{equation}
			\begin{aligned}
				R_L	&= \prod_0^{D} (1-p_o(x_d(d))))^{\dif d}\\
						 &= (1-p_o)^{D} \quad \text{ in case of homogeneous field,}
			\end{aligned}
			\label{eq:HeidenR_L}
			\end{equation}
			where $D=\int_0^T |\dot{x}(t)|\dif t$ is the total distance crossed by the robot and $x_d(\cdot)$ is the position of the robot as a function of the traveled distance.
			Using Equation (\ref{eq:HeidenR_L}), the~probability of collision does not depend on the tessellation size or the speed of the vehicle.
			The main drawback is that there is no natural reason to use the concept of product integrals in Bayesian occupancy grids, as~it is here merely a tool to make the probability constant.
			Furthermore, the~robot is considered to be reduced to a point.
			The well-known solution to this problem is to inflate the obstacles, at~the cost of assuming that the robot is round. 
			As the Lambda Field takes into account the size of the robot, we propose an improvement of their theory to take into account the robot's width $W$.
                        Instead of only integrating over the robot line path, we also integrate over the entire width of the robot (i.e., the size of the front of the robot) for each position $x_d(d)$.
			Under this consideration, the~reachability equation becomes
\begin{equation}
				R_L = \prod_0^D\prod_{-W/2}^{W/2} \left(1-p_o(x(d,w))\right)^{\dif w \dif d} ,
				\label{eq:HeidenR_L_improved}
			\end{equation}
                        where $x(d,w)$ is a point of the robot parametrized as the distance that the robot has traveled $d$ and the distance from the center of the robot head in its width direction $w$.
			We can see that for the special case $W=1$ and $p_o(x(d,w))$ constant for $w\in[-W/2, W/2]$, we fall back on Equation (\ref{eq:HeidenR_L}).
			Assuming that the robot fully crosses the cells it encounters, we can develop a more convenient formulation for calculations.
                        If the robot fully crosses the $N$ cells $\mathcal{C}=\{c_i\}_{0:N-1}$ of size $\SI[parse-numbers=false]{S \times S}{\m\squared}$ and probability of occupancy $p_{oi}$, Equation (\ref{eq:HeidenR_L_improved}) can be rearranged to give
\begin{equation}
			\begin{aligned}
				R_L &= \prod_{i=0}^{N-1} \prod_{x=0}^{S} \prod_{y=0}^{S} (1-p_{oi})^{\dif x \dif y} \\
					 &= \prod_{i=0}^{N-1} (1-p_{oi})^{\area},
			\end{aligned}
			\label{eq:HeidenR_L_improved_cells}
			\end{equation}
			where $\area=S^2$ is the area of each~cell.

                        From there, the~improvement in Equation (\ref{eq:HeidenR_L_improved_cells}) of the theory of \citet{Heiden2017} can be linked to the theory of the Lambda Field.
                        Indeed, for~a path crossing the cells $\mathcal{C}=\{c_i\}_{0:N-1}$ in a Lambda Field, the~probability of not colliding during the traversal is computed as
\begin{align}
					1-\mathbb{P}({\mathtt{coll}}) &= \expp{-\area\sum_{c_i\in\mathcal{C}}\lambda_i} \nonumber\\
			\label{eq:relationReachabilityLambda}
                                                                      &= \prod_{i=0}^{N-1}\expp{-\lambda_i}^{\area} \\
                                                                      &= \prod_{i=0}^{N-1}(1-p_{oi})^{\area} \text{ with } p_{oi}=1-\expp{-\lambda_i} . \nonumber
			\end{align}

                        Hence, the~probability of occupancy $p_{oi}$ of a cell in~\cite{Heiden2017} is the probability of colliding in the cell $c_i$ of area $\SI{1}{\m\squared}$ in a Lambda Field. %
			Therefore, with~our improvement given by Equation (\ref{eq:HeidenR_L_improved_cells}), the~theory of \citet{Heiden2017} is then a special case of our framework, where the risk function $r(\cdot)$ is set to $1$ and the area of the cells is assumed to be equal to $1$.
			Compared to~\cite{Heiden2017}, we propose a more meaningful approach, where the theory provides a way to assess risk that is not restricted to be the probability of~collision.

\section{Validation} \label{sec:experiment}
	\subsection{Setup}
\begin{figure}[H]
		
		\includegraphics[width=.6\linewidth]{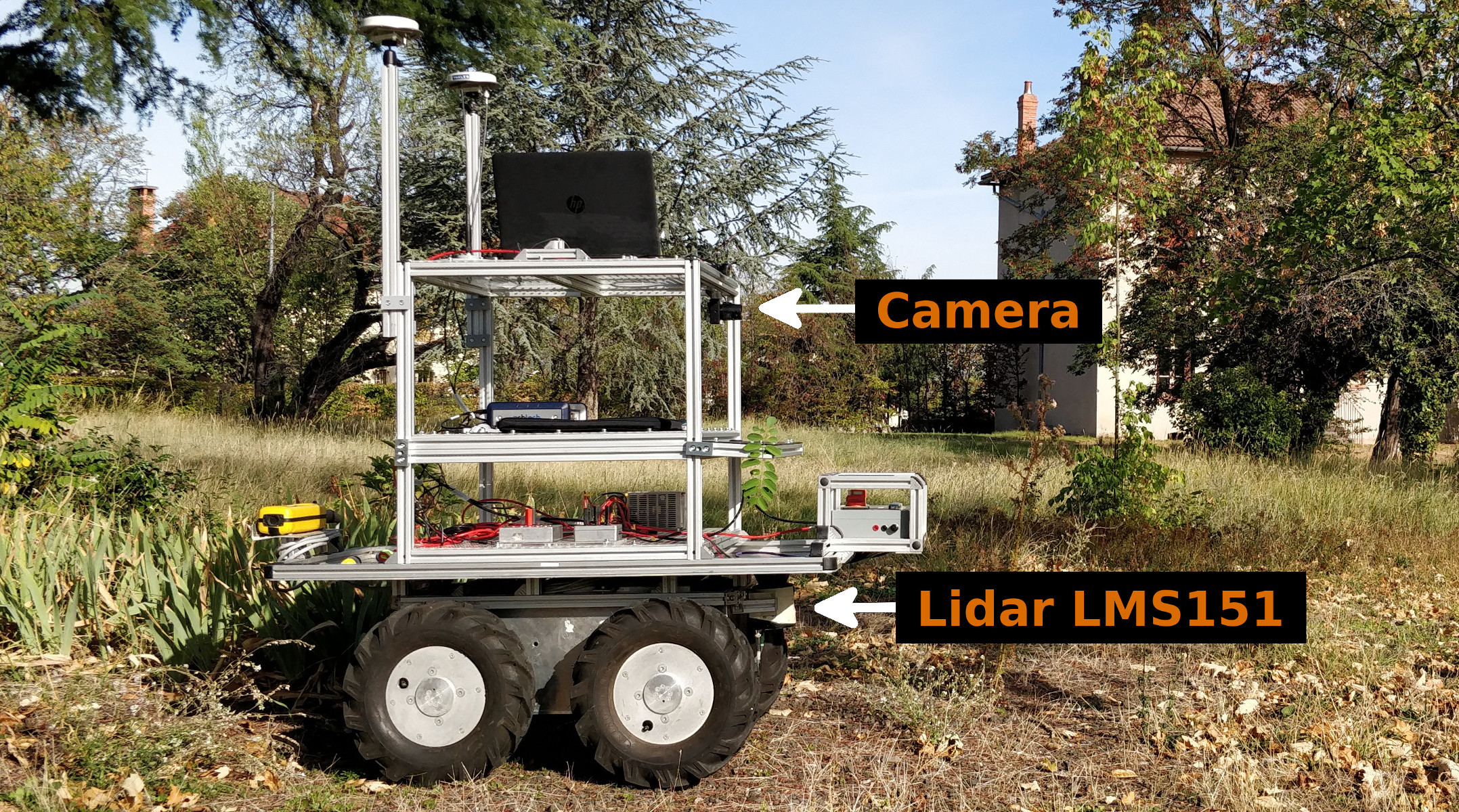}
		\caption{Robot used in the experimentations. It is equipped with a Sick LMS151 lidar and a~camera.}
		\label{fig:robot_setup}
	\end{figure}

	We implemented our framework in a robot equipped with an LMS151 lidar and a camera, as~shown in Figure \ref{fig:robot_setup}.
	Since the robot has four-wheel steering, it was not  impacted much by slipping and skidding, and the odometry was sufficient to estimate the robot~displacements.

	{For every new lidar scan, the~displacement between the current and previous position was estimated, and the map was updated.
        The Lambda Field was estimated at each iteration using} Equation (\ref{eq:lambda}).
	Moreover, the~map was centered on the robot.
	We chose not to rotate the map, but to rotate the robot instead to nullify the errors coming from the rotation.
	Indeed, a~straight wall was quickly distorted after a few rotations because of the tessellation of the map.
	We also had to keep a global offset of the map, as otherwise, small displacements were not taken into account.
	Without this offset, the map was not precisely updated, as any displacement below half the cell size was discarded.
	The mass of the robot was also set to $m_R = \SI{50}{\kg}$.
	For safety purposes, the~maximum speed of the robot was set to $\SI{0.5}{\m\per\s}$, and the maximum acceleration to $\SI{0.05}{\m\per\s\squared}$.
	The parameters used for the confidence intervals of the mapping were $p^m=0.9999$ and $p^h=0.99$ for all cell measurements, where the cells had a size of $\SI[parse-numbers=false]{0.1\times0.1}{\m}$.
	The value of $p^m$ was intentionally very high, as it was nearly impossible for a lidar beam to go through obstacles.
	The normals of the obstacles were estimated using the method developed in~\cite{Senanayake2018}.
	At each lidar scan, the~normals of the points were estimated by using their nearest neighbors, and the normal of each underlying cell $c_i$ was updated as follows:
\begin{equation}
		\bar{\theta} = %
		\begin{cases}
			\arctan(\bar{S}/\bar{C}) &\text{if } \bar{C} \geq 0 \\
			\arctan(\bar{S}/\bar{C})+\pi &\text{otherwise,}
		\end{cases}
	\end{equation}
	with
\begin{equation}
          \bar{C} = \sum_{k=0}^{N_i-1} \cos(\theta_k) \quad \text{and} \quad%
          \bar{S} = \sum_{k=0}^{N_i-1} \sin(\theta_k),
	\end{equation}
	where $N_i$ is the number of normal measurements $\theta_k$ for the cell $c_i$.

	\subsection{Comparison with the Bayesian Occupancy~Grid}
	In order to demonstrate the discrepancies between the Bayesian occupancy grid and the Lambda Fields, we theoretically investigated the key differences between the two frameworks, and then,  in real-world experiments, showed their consequences on the quality of the~maps.
	
	First, we investigated the convergence of the occupancy of a single cell.
	In the context of unstructured environments, it is very common to have cells that are only partially occupied.
	This can come from either very thin objects, such as tall grass or crops, or~from obstacles that do not reflect the laser beam well, such as a dense bush.
	In both cases, the~cell will be measured as both `hit' and `miss', as the beam can cross or hit the obstacles in the cell.
	Using the theory presented in~\cite{Thrun2005} to construct the Bayesian occupancy grid, we used the log odds representation of occupancy.
	Assuming that a cell is measured $N$ times and is filled at a ratio of $r\in[0,1]$ (1 is completely filled, 0 is completely empty), the~cell will be measured `hit' $rN$ times and `miss' $(1-r)N$ times.
	The Bayesian occupancy grid estimates the occupancy probability of the cell as
\begin{equation}
		\P{\mathtt{occ}_b} = 1 - \frac{1}{1 + \expp{rNl_o + (1-r)Nl_f}},
	\end{equation}
	where $l_o$ is the log odds representation of the probability that the cell is occupied given a `hit' measurement, whereas $l_f$ is the log odds representation that the cell is occupied given a `miss' measurement (i.e., it informs that the cell is free).
	These quantities are computed as
\begin{equation}
		\begin{aligned}
			l_o &= \ln\left(\frac{\P{\mathtt{occ} | z = \mathtt{hit}}}{1-\P{\mathtt{occ} | z = \mathtt{hit}}}\right), \text{ and}\\
			l_f &= \ln\left(\frac{\P{\mathtt{occ} | z = \mathtt{miss}}}{1-\P{\mathtt{occ} | z = \mathtt{miss}}}\right),
		\end{aligned}
	\end{equation}
	with $z$ being the measurement of the sensor that is either `hit' or `miss'.
	Substituting the definition of the log odds representation with its expression, we have
\begin{equation}
		\begin{aligned}
                  \P{\mathtt{occ}_b} &= 1 - \frac{1}{1 + \left[\left(\frac{\P{\mathtt{occ} | z = \mathtt{hit}}}{1-\P{\mathtt{occ} | z = \mathtt{hit}}}\right)^r\left(\frac{\P{\mathtt{occ} | z = \mathtt{miss}}}{1-\P{\mathtt{occ} | z = \mathtt{miss}}}\right)^{1-r}\right]^N} \\
                        &\triangleq 1 - \frac{1}{1 + \left[O_o^r \cdot O_f^{1-r}\right]^N}, \\
		\end{aligned}
	\end{equation}
        \sloppy %
	where $O_o,O_f \in \mathbb{R}_{\geq 0}$ are defined as the odds of, respectively, $\P{\mathtt{occ} | z = \mathtt{hit}}$ and $\P{\mathtt{occ} | z = \mathtt{miss}}$.
	Taking the limit $N\rightarrow \infty$, we have
\begin{equation}
		\lim_{N\rightarrow \infty} \P{\mathtt{occ}_b} = %
		\begin{cases}
			1 &\text{ if } O_o^r \cdot O_f^{1-r} < 1 \\
			0.5 &\text{ if } O_o^r \cdot O_f^{1-r} = 1 \\
			0 &\text{ if } O_o^r \cdot O_f^{1-r} > 1. \\
		\end{cases}
		\label{eq:lim_bayes}
	\end{equation}
	
	Therefore, we see that the Bayesian occupancy grid will always converge to an extremum (apart from the special case where $O_o^r \cdot O_f^{1-r} = 1$, meaning that the measurements do not provide information about the occupancy).
	On the contrary, the~Lambda Field does not converge to an extremum.
	Indeed, putting the estimation of lambda given by Equation~(\ref{eq:lambda}) into the probability of collision of Equation (\ref{eq:collision_proba}), we have
\begin{equation}
		\begin{aligned}
			\P{\mathtt{occ}_\lambda} &= %
									  1 - \expp{-\area \cdot \frac{1}{e}\ln\left(1+ \frac{Nr}{N(1-r)}\right)} \\
									  &= 1 - \left(1 + \frac{r}{1-r}\right)^{-\frac{\area}{e}}.
		\end{aligned}
	\end{equation}
	
\begin{figure}[H]
			
			\includegraphics[width=.9\linewidth]{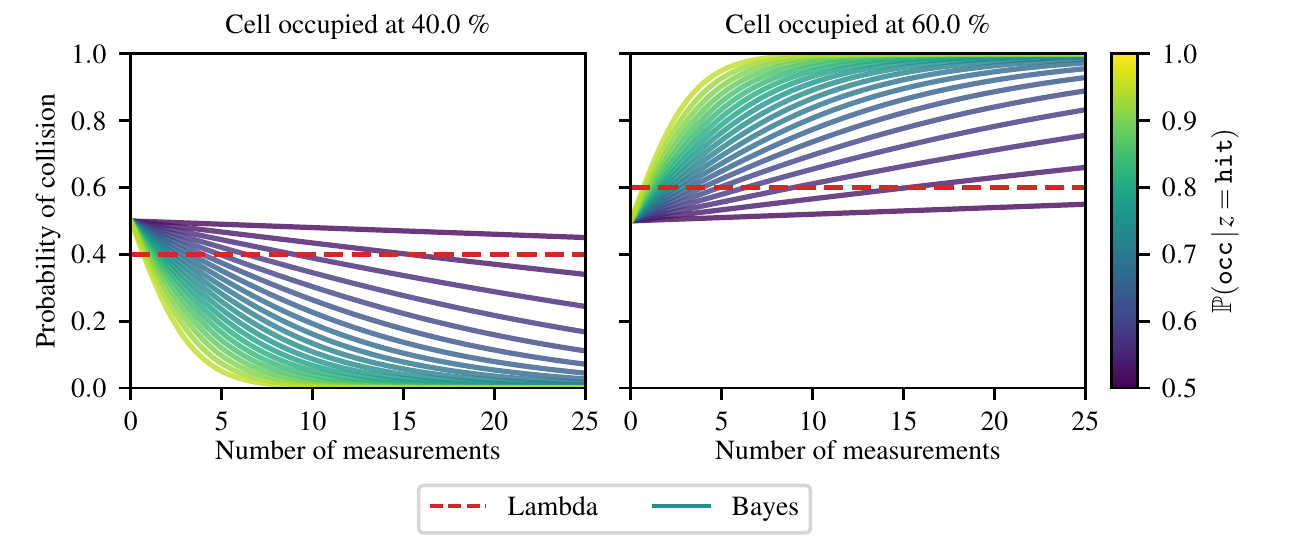}
			\caption{Convergence of the collision probability for a given cell occupied at 40\% and 60\%. 
					 The Bayesian occupancy grid will quickly converge to a probability of occupancy of either 0 or 1, whereas our framework stays at the true occupancy value of the cell.}
			\label{fig:convergence_lambda_bayes}
	\end{figure}

	In the case where the lidar error region $e$ is equal to the area of the cells $\area$, meaning that we are sure that the collision comes from this cell, the~equation simplifies to
\begin{equation}
		\begin{aligned}
			\P{\mathtt{occ}_\lambda} &= 1 - \frac{1}{1 + \frac{r}{1-r}} \\
									  &= r,
		\end{aligned}
	\end{equation}
	meaning that the probability of collision is purely the ratio of occupancy of the cell and does not depend on the number of measurements $N$.
	Figure \ref{fig:convergence_lambda_bayes} shows the convergence of the Bayesian occupancy grids and the Lambda Field.
	In order to ease the reading, the~amount of information for a `hit' measurement is the same as for a `miss' measurement of the cell, meaning that $\P{\mathtt{occ} | z = \mathtt{hit}} = 1-\P{\mathtt{occ} | z = \mathtt{miss}}$ and thus $l_o = -l_f$.

	The behavior still remains the same without this assumption, as shown by \mbox{Equation~(\ref{eq:lim_bayes}}).
	The Lambda Field estimation stays at the true occupancy value of the cell, whereas the occupancy probability of the Bayesian occupancy grid converges to either 0 or 1 depending on the occupancy.
	As expected, the~higher the confidence of the sensor's measurement, the~faster the probability converges.
	This behavior can be hazardous in an unstructured environment, as~a cell filled at 49\% will converge to a probability of occupancy of $0$ after a few seconds.

\begin{figure}[H]
		
		\includegraphics[width=.9\linewidth]{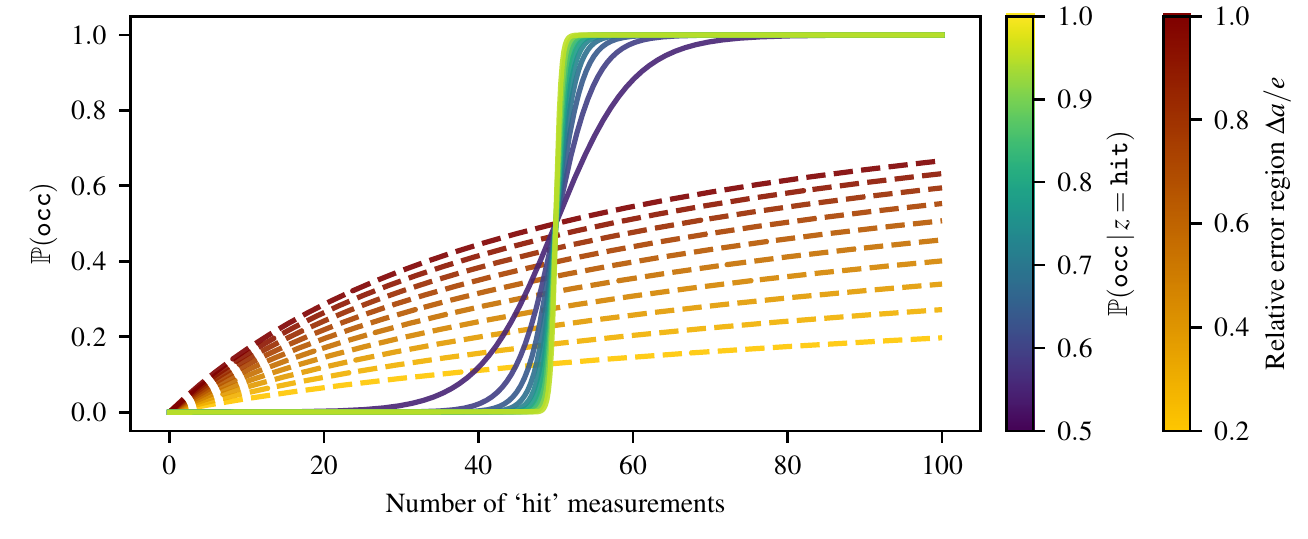}
		\caption{Evolution of the collision probability of a fully occupied cell that has been wrongly measured as empty 50 times before. As~theoretically shown, the~Lambda Field manages to recover more quickly from the wrong estimation, whereas the Bayesian occupancy grid converges faster to the true state of the cell after 50 `hit'~measurements.}
		\label{fig:convergence_recovery}
	\end{figure}

	We also consider the case where an obstacle is wrongly undetected. 
	This situation can happen when measuring sparse or unstructured obstacles.
	For instance, a~wire fence can either stop or let through the laser beams depending on the position of the robot.
	Thus, we look at the speed at which the Bayesian occupancy grid and the Lambda Field can recover the true state of an obstacle wrongly labeled as free.
	For a single cell measured $m$ times as `miss', we assume that $m$ is large enough such that the Bayesian occupancy grids converge to the probability of occupancy $\P{\mathtt{occ}}=0$.
	Then, assuming that after $m$ measurements, the~robot starts to measure the obstacle as `hit', we can approximate the probability of occupancy around the number of `hit' measurements $h\approx 0$ using first-order Taylor expansion, as~\begin{equation}
	\begin{aligned}
		\P{\mathtt{occ_b}} &= 1 - \frac{1}{1+\expp{ml_f + hl_o}} \\ 
						   &\approx \frac{l_o\expp{ml_f}}{\left(\expp{ml_f}+1\right)^2}\cdot h \\
						   &\approx \frac{l_o}{2\left(\cosh(ml_f)+1\right)}\cdot h,
	\end{aligned}
	\end{equation}
	meaning that the recovery rate of the Bayesian occupancy grid vanishes exponentially as a function of the number of times $m$ the cell has been wrongfully measured.
	In the case of the Lambda Field, we have
	
\begin{equation}
	\begin{aligned}
            \P{\mathtt{occ}_\lambda} &= 1 - \expp{-\frac{\area}{e}\ln\left(1+\frac{h}{m}\right)} \\ 
								 &= 1 - \frac{1}{\left(1+h/m\right)^{\area/e}} \\
								 &\approx \frac{\area/e}{m}\cdot h,
	\end{aligned}
	\end{equation}
	indicating that the recovery rate vanishes linearly as a function of the number of times $m$ the cell has been wrongfully measured.
	Figure~\ref{fig:convergence_recovery} shows the convergence curves towards the true state of the cell (i.e., 100\% filled) for different values of confidence of the sensor measurements.
	The cell was  measured 50 times as `miss' before (e.g., the robot stayed still during \SI{2}{\s} for a \SI{25}{\Hz} lidar such as the Sick LMS151).
	Then, the~robot changed position, allowing the lidar beams to hit the obstacle in the cell, leading to `hit' measurements in the cell.
	As expected, the~higher the confidence on the sensor (i.e., smaller error region for the Lambda Field and higher probability for the Bayesian occupancy grid), the~faster the recovery speed is.
	However, the~Lambda Field allows  better recovery at the beginning by growing faster, whereas the Bayesian occupancy grid prefers to quickly converge to the `occupied' state after 50 `hit' measurements.
	The Lambda Field does, however, take more time to converge toward a full occupancy of the cell, as it still takes into account the previous wrong `miss' measurements.
	Indeed, as~shown in Figure~\ref{fig:convergence_lambda_bayes}, the~Bayesian occupancy grid only converges to zero or one.
	Therefore, as~soon as the `hit' measurements become predominant over the wrong `miss' measurements, the~framework quickly converges to 1.
	Both frameworks can have their convergence speed shortened by applying a threshold on the probability of occupancy and the lambda (i.e., cannot go above or below certain values).
	However, this enhancement does not modify the previous analysis, as it only bounds the vanishing of the recovery~rate.

\begin{figure}[H]
\widefigure
	\begin{tabular}{ccc}
\includegraphics[width=0.3\linewidth]{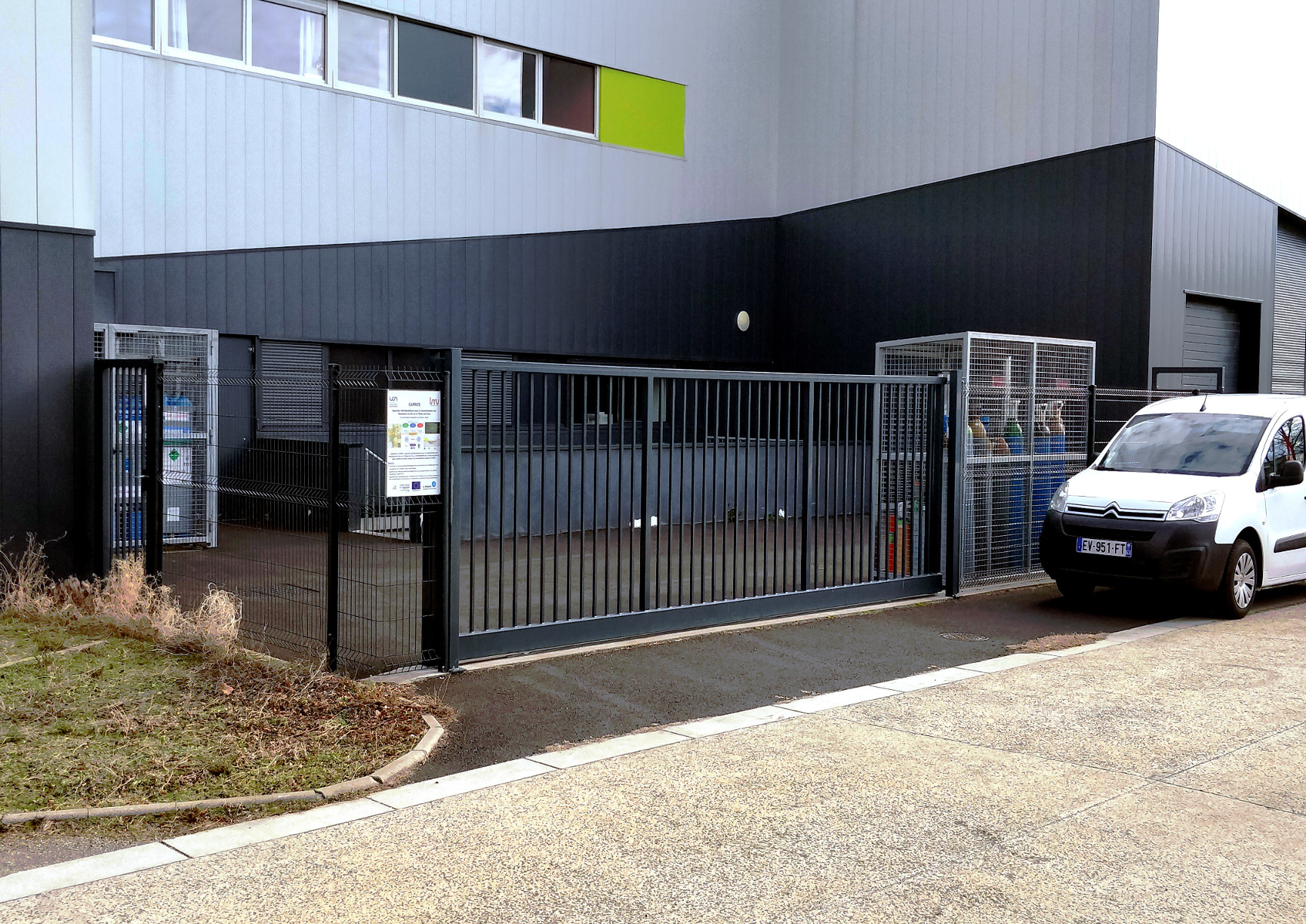}&
\includegraphics[width=0.3\linewidth]{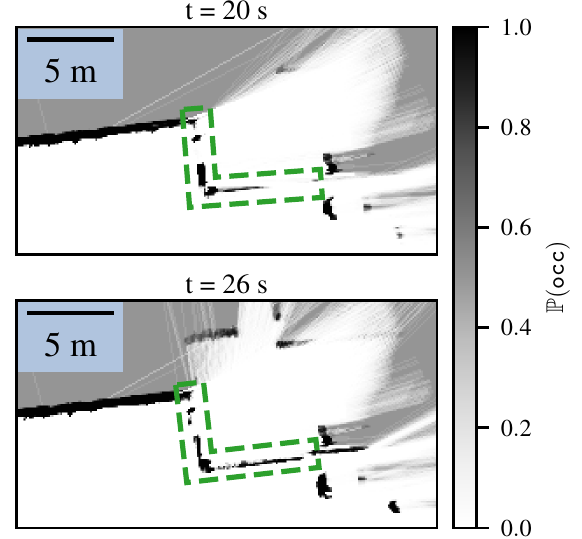}&
	\includegraphics[width=0.3\linewidth]{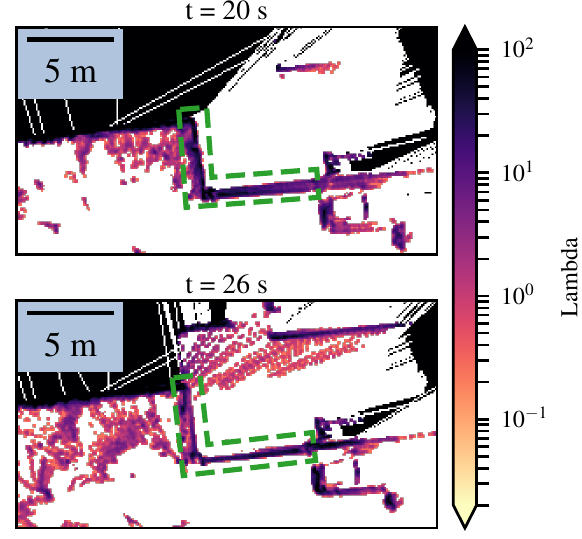}\\
({\bf a})&({\bf b})&({\bf c})\\
\end{tabular}

			\caption{Mapping
 of a wire fence where, depending on the robot pose, the~laser beams can either collide or go through the fence  (outlined in dashed green in the maps). 
(\textbf{a}): Picture of the mapped environment.
 (\textbf{b}) Bayesian occupancy grid of the wire fence. 
 (\textbf{c}) Lambda Field of the wire fence. One can note that the Bayesian occupancy grid did not have the time to converge at $t=\SI{20}{\s}$ because of the wrong `miss' measurements at the beginning of the~experiment.}
		\label{fig:wireFence}
	\end{figure}

	In order to demonstrate these considerations in real-world conditions, we mapped a small zone consisting of a black wire fence where lidar beams could easily get through, as~shown in Figure~\ref{fig:wireFence}.
	At first, the~position of the robot led the laser beams to cross the fence without collision, yielding both the Bayesian occupancy grid and the Lambda Field to converge to a false state.
	After a few seconds, the~robot turned in front of the fence, leading more lasers beams to actually collide with the obstacle.
	In this configuration, the~two aforementioned differences between the Lambda Field and Bayesian occupancy grid were involved.
	On the one hand, the~laser beams still had a chance to go through the fence, leading  cells that were not completely filled to wrongly converge for the Bayesian occupancy grid. %
	If the lasers collided with the cell less than 50\% of the time, the~Bayesian occupancy grid would wrongly converge to 0.
	On the other hand, because~of the wrong `miss' measurements at the beginning, the~Bayesian occupancy grid would struggle more to recover the true state.
	\textls[-5]{In order to measure the quality of the map, we manually labeled the fence at each iteration and used patches of $4\times 4$ cells running along the fence.
        The inspected zone is shown in Figure~\ref{fig:wireFence} in dashed green.}

\begin{figure}[H] 
\widefigure
	\begin{tabular}{cc}
\includegraphics[width=.27\linewidth]{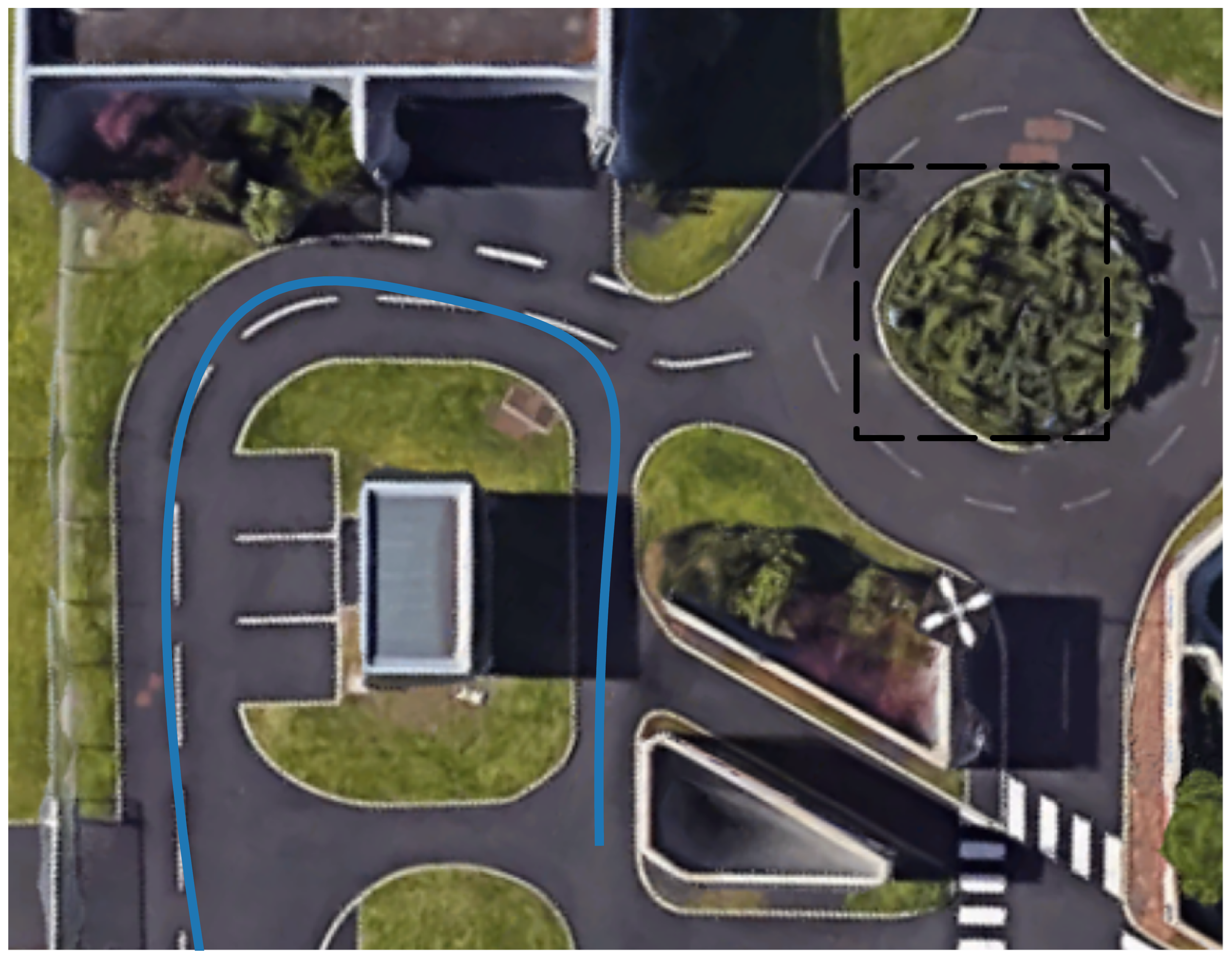}
&\includegraphics[width=.62\linewidth]{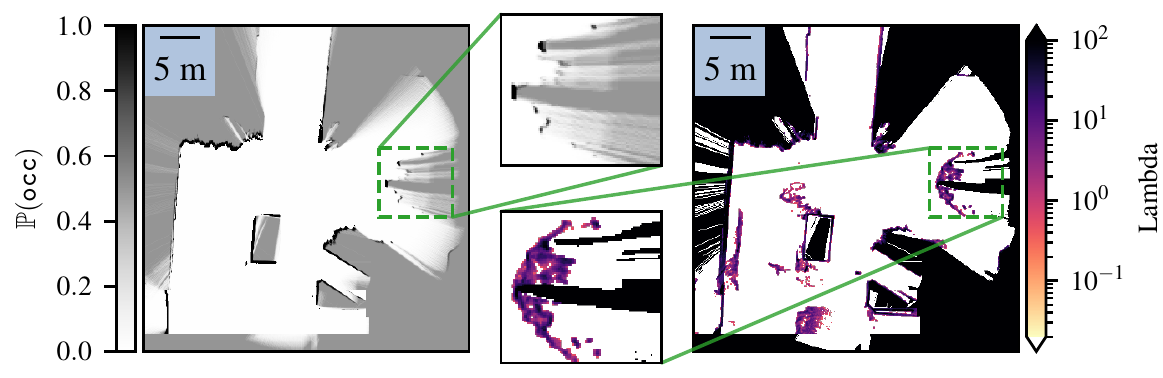}\\
({\bf a})&({\bf b})\\
\end{tabular}

\caption{(\textbf{a}) Aerial view of the mapped environment, with~the robot path in blue and the roundabout in dashed black. (\textbf{b}) \textit{Left:} Bayesian occupancy grid; \textit{Right:} Lambda Field. The~Lambda Field is better suited for storing the occupancy of unstructured obstacles where the Bayesian occupancy grid may over-converge, especially for the~roundabout.}
		\label{fig:compareMaps}
	\end{figure}
\begin{figure}[H] 
\widefigure
	\begin{tabular}{ccc}
\includegraphics[width=.25\linewidth]{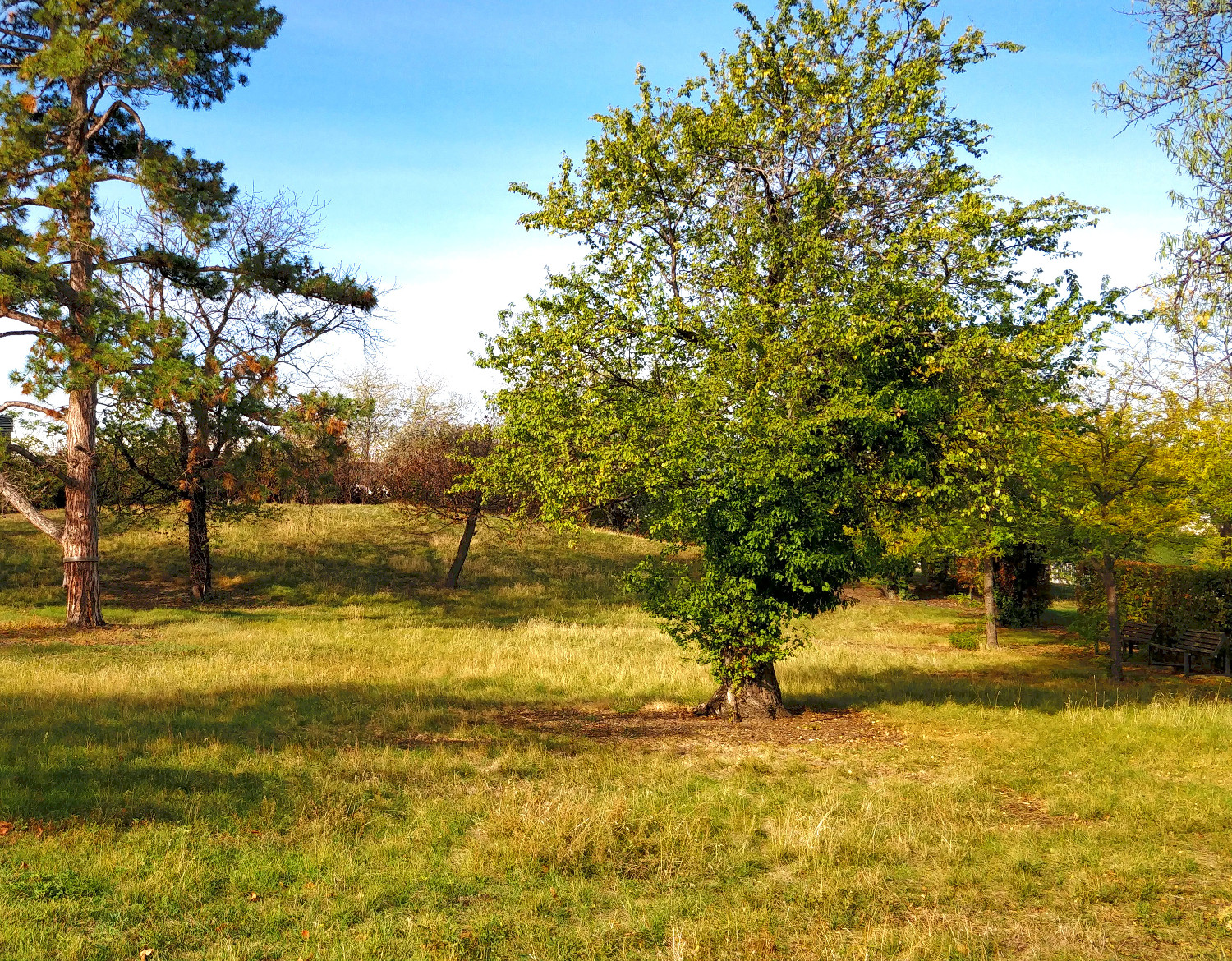}
&\includegraphics[width=.32\linewidth]{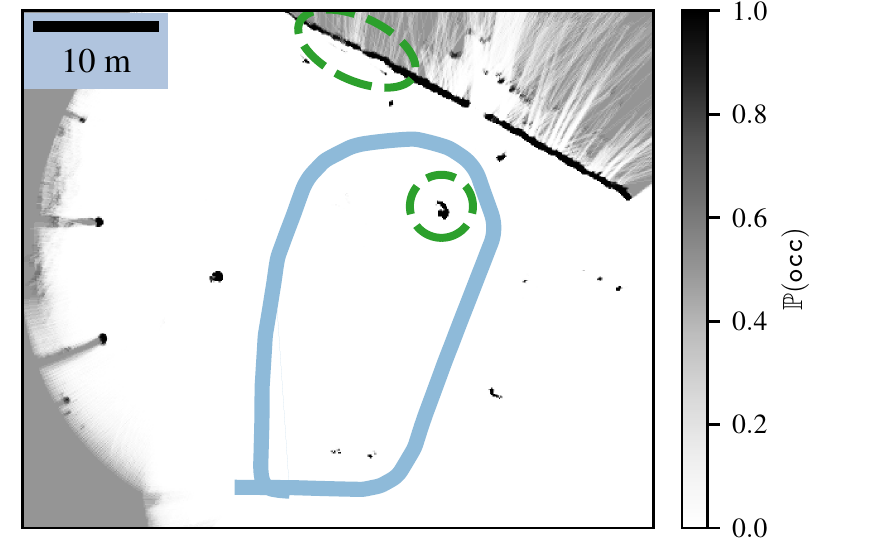}
&	\includegraphics[width=.32\linewidth]{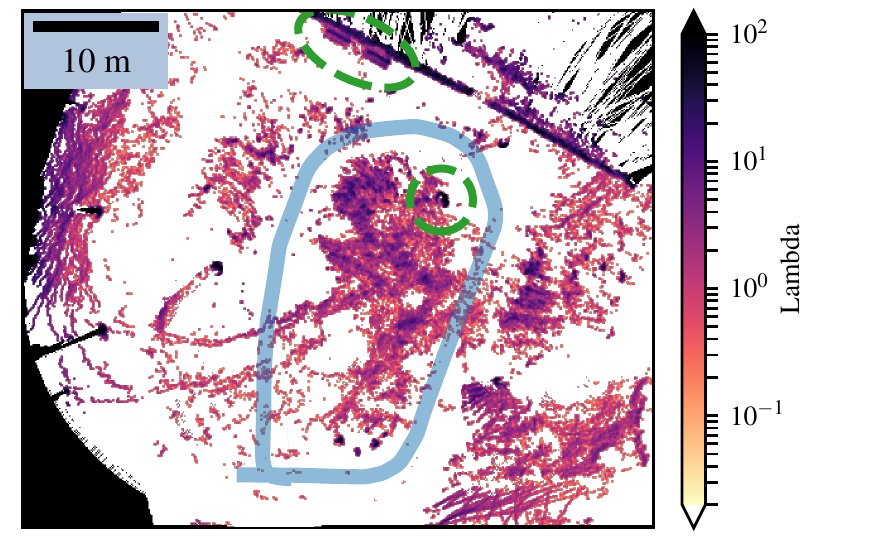}\\
({\bf a})&({\bf b})&({\bf c})\\
\end{tabular}

\caption{Mapping of an unstructured zone. 
                (\textbf{a}) Picture of the environment.
              (\textbf{b}) Bayesian occupancy grid of the environment. 
              (\textbf{c})~Lambda Field of the environment. The~robot, with~its path in light blue, went around the nearest tree (circled in green) before~going back to its initial position. Because~of the tall grass, the~Lambda Field stored a lot of obstacles during the traversal, while the Bayesian occupancy grid discarded the vast majority of them. The~hedge and the tree trunk (circled in green) are still visible in both maps, as they are the only structured obstacles, whereas more unstructured obstacles, such as benches on the top of the map (circled in green) or bushes, were  discarded by the Bayesian occupancy~grid.}
		\label{fig:MappingIrstea}
	\end{figure}
\begin{figure}[H] 
\widefigure
	\begin{tabular}{ccc}
\includegraphics[width=0.3\linewidth]{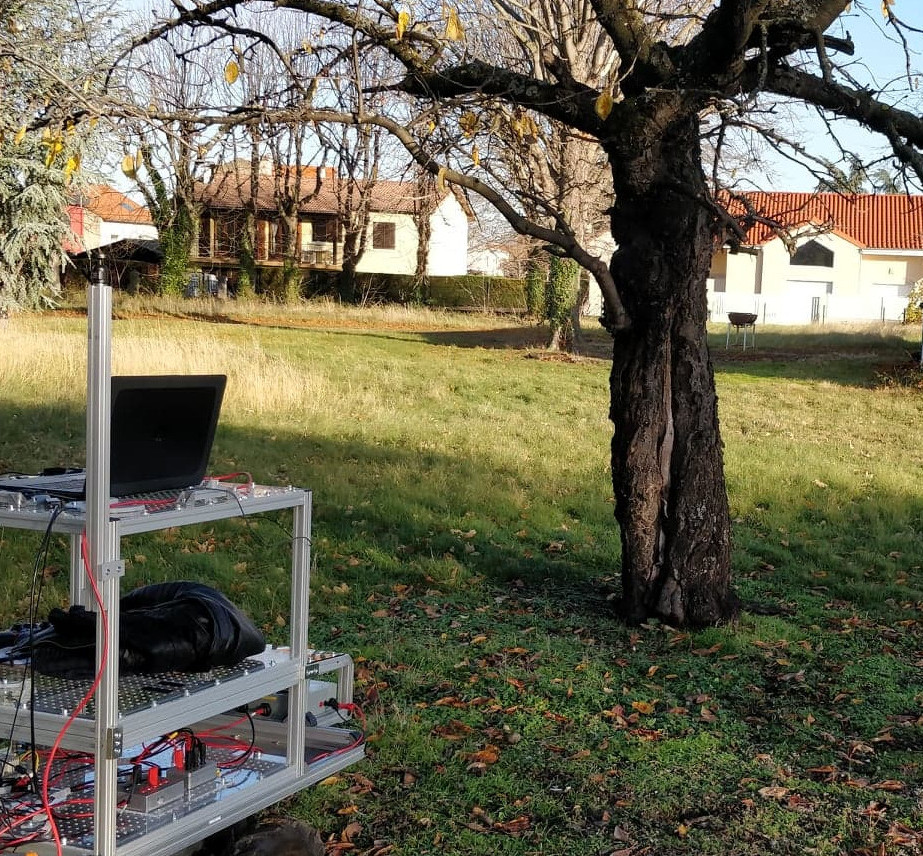}&   \includegraphics[width=0.3\linewidth]{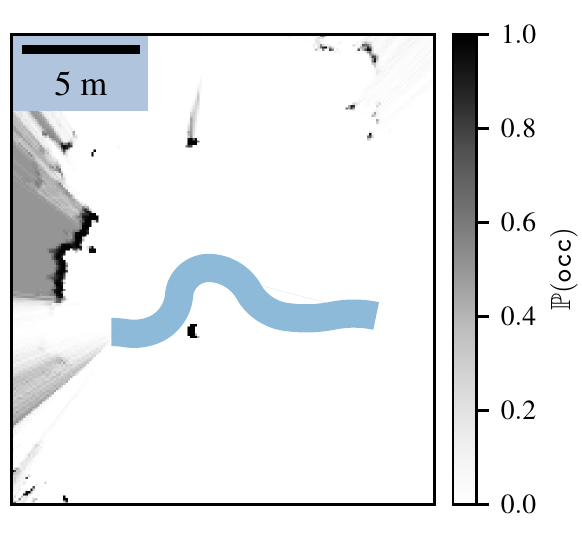} &  \includegraphics[width=.3\linewidth]{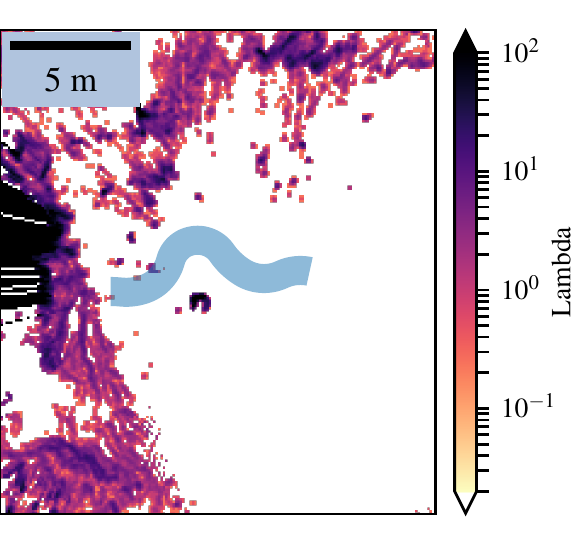} \\
({\bf a})& 
({\bf b})&
({\bf c})\\
\end{tabular}

	                \caption{The
 robot had to avoid a tree that was on its path. 
 (\textbf{a}) Picture of the environment.
 (\textbf{b}) Bayesian occupancy grid with the path that the robot  took in light blue. (\textbf{c}) Lambda Field with the path the robot  took in~light blue.}
		\label{fig:basicPathPlanning}
	\end{figure}

\end{paracol}
\nointerlineskip
\begin{paracol}{2}
\switchcolumn

	Using patches instead of directly analyzing the cells removes the noise due to the manual labeling and the noisy odometry of the robot.
	Using these patches, we evaluated the recall of the detection, computed as the sum of the probabilities of not colliding in each patch $p_k$ over the number $P$ of patches (i.e.,~the proportion of wrongly detected patches) as
\begin{equation}
		\begin{aligned}
			\text{Recall}_b &= \frac{1}{P}\sum_{k=0}^{P-1}\prod_{c_i\in p_k}\left(1-\P{c_i = \mathtt{occ}}\right), \\
			\text{Recall}_\lambda &= \frac{1}{P}\sum_{k=0}^{P-1} \expp{-\area\sum_{c_i\in p_k}\lambda_i}.
		\end{aligned}
	\end{equation}
	In addition, the~recall  can be seen as the mean of the probabilities of not colliding in the patches.
        Thus, we also computed the associated standard deviation of the patches.
	\mbox{Figure~\ref{fig:convergence_wireFence}} depicts the recall for the Lambda Field and Bayesian grid, as well as the distribution at two sigmas (approximately 95\%) of the probability of not colliding in the~patches.
	\begin{figure}[H]
		
		\includegraphics[width=.9\linewidth]{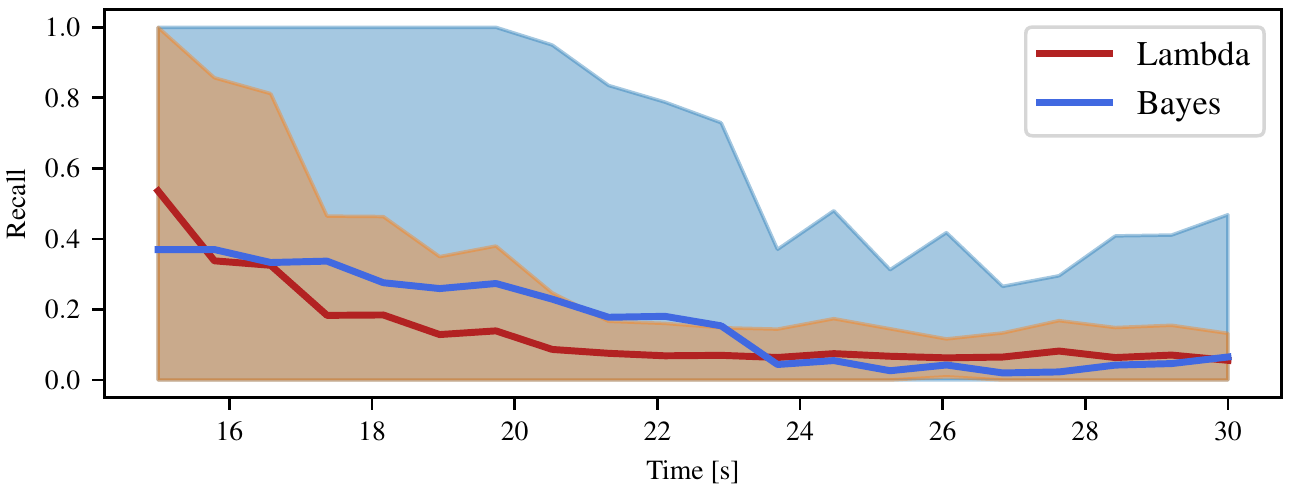}
		\vskip-.5em
		\caption{Mapping error of the wire fence for the Bayesian occupancy grid and the Lambda Field. The~error is defined as the ratio of free space over the whole space that represents the obstacle. 
		As expected, the~Lambda Field converges more quickly to a low error, whereas the Bayesian occupancy grid needs more time to assess its occupancy. 
	}
		\label{fig:convergence_wireFence}
	\end{figure}

	As theoretically expected, the~Lambda Field recovered more quickly from the wrong `free' state of the wire fence.
	At $t=\SI{20}{\s}$, the~Lambda Field converged to the state where the whole fence was considered as an obstacle, whereas the Bayesian occupancy grid was still converging and had more than 20\% of the fence that was considered as free.
	At $t\approx \SI{23}{\s}$, the~robot changed position such that parts of the fence were not visible to the lidar sensor.
	As such, the~Lambda Field started to converge toward a lower lambda that was mainly seen on the left vertical wall.
	Although the recall was lower for the Bayesian occupancy grid, one can still see holes in the fence at $t=\SI{26}{\s}$, while the Lambda Field had the whole fence mapped.
        Indeed, the~distribution at  {95}\% of the probabilities of the patches for the Lambda Field was always contained in the one of the Bayesian occupancy grid.
	This means that although the recall was lower, the~Bayesian occupancy grid had patches that were poorly represented compared to the Lambda Field.
	One can also note that the car on the right of the map was better represented in the Lambda Field than in the Bayesian occupancy grid.
        The low-lambda obstacles on the left of the Lambda Field correspond to tall grass and unstructured, sparse vegetation.
	%

        Next, we show the effectiveness of our framework in mapping large environments.
	To do so, we implemented a simple robot follower scenario in an urban-like environment.
	The robot had to follow a pedestrian while keeping the risk of the chosen path below $\SI{5}{\kg\m\per\s}$.
	While following the pedestrian, the~robot created a Lambda Field as well as a Bayesian occupancy grid of the environment, as~shown in Figure~\ref{fig:compareMaps}.
	The maps were globally alike, except for the unstructured obstacles, which were, in this case, the bushes in the roundabout, as well as tall grass around the pavement.
	As the Bayesian occupancy grid needed to converge to either the `occupied' or `free' state, a~lot of information about the occupancy of the roundabout was discarded.
	Furthermore, the~robot was not able to see the entire obstacle at first, leading the frameworks to wrongly converge.
	As shown in the previous section, the~Lambda Field recovered faster in these situations, leading to a more precise map.
	As such, most of the information was preserved in the Lambda Field, and the global shape of the roundabout was more easily recognizable.
	The other disparities between the two maps also came from unstructured obstacles, which were small trees and tall grass.
	\end{paracol}
\nointerlineskip
\begin{paracol}{2}
\switchcolumn

	Finally, we also mapped an unstructured environment, where the resulting maps are shown in Figure~\ref{fig:MappingIrstea} and the environment is depicted in Figure~\ref{fig:MappingIrstea}a.
	The environment consisted of several trees with a lot of tall grass disrupting lidar measurements.
	The robot went around the tree in the center of the picture while navigating in the grass.
	However, due to the grass and the wind, the~lidar returned many measurements corresponding to the grass.
	Whereas the Bayesian occupancy grid only kept the hedge and the main trees, the~Lambda Field kept more information, such as the wooden benches on the top of the map or the tall grass.
	This behavior was caused by the necessity of the Bayesian occupancy grid to always converge to an extremum, leading it to discard a lot of information that could be critical.
	For instance, in~the case of agricultural robotics, keeping the crops on the map is essential for not rolling over them.
\end{paracol}
\nointerlineskip
\begin{paracol}{2}
\switchcolumn

	\subsection{Basic Path~Planning}
Here, we demonstrate  that our framework can be used to perform classical path planning.
As shown in Figure~\ref{fig:basicPathPlanning}a, the~robot had to go around a tree to reach the goal that was set behind.
To do so, we implemented the path-planning algorithm of~\cite{Gerkey2008}.
Every 3~s, we sampled feasible commands for the robot, that is, a velocity and steering angle, and~chose the best one.
The best path (i.e., command applied for \SI{3}{\s}) was the one that stayed below a risk threshold and led the robot as close as possible to the target, which was, in this case, behind the tree.
The chosen path also required  an upper risk (i.e., the risk computed using the upper bound of the lambdas) below a certain risk threshold.
For each feasible command, the~$N$ cells crossing the path induced by the command applied for 8 s were extracted and the risks were computed.
Estimating the risk of a longer time than the one applied by the command avoided the robot choosing paths that led to a dead end.
Indeed, if~the risk was computed only for the time of the command, the~applied command might have led the robot to be right in front of a wall in a configuration where it was impossible to escape.
In the case where no command met the criteria, the~robot stopped. 
This could happen when the robot was at high speed; because of the limited deceleration, all of the high-speed commands led to a risk higher than the maximum  that was allowed. %
Then, the~robot completely stopped before continuing its course, as it could now sample low-speed~commands.
	%
	%
	%
	%
	%
	%


We also implemented the reachability metric of~\cite{Heiden2017} with our improvement for handling the robot size.
As converting the Lambda Field into an occupancy grid using \mbox{Equation (\ref{eq:relationReachabilityLambda})} would lead to computing the risk $r(\cdot)=1$ and using our theory, we used the Bayesian occupancy grid directly computed from lidar measurements.
Using the same method for path planning, we sampled paths and chose the one that led the closest to the goal, where a path was considered safe if its reachability $R_L$ was above a certain threshold $1-\epsilon$, with~$\epsilon\in(0,1)$.
The threshold $\epsilon$ was set to $0.1$ during our experiments.
For each command applied, 300 samples were~evaluated.	

\end{paracol}
\nointerlineskip
\begin{paracol}{2}
\switchcolumn

\vspace{-10pt}{}
\begin{figure}[H] 
		
		\includegraphics[width=\linewidth]{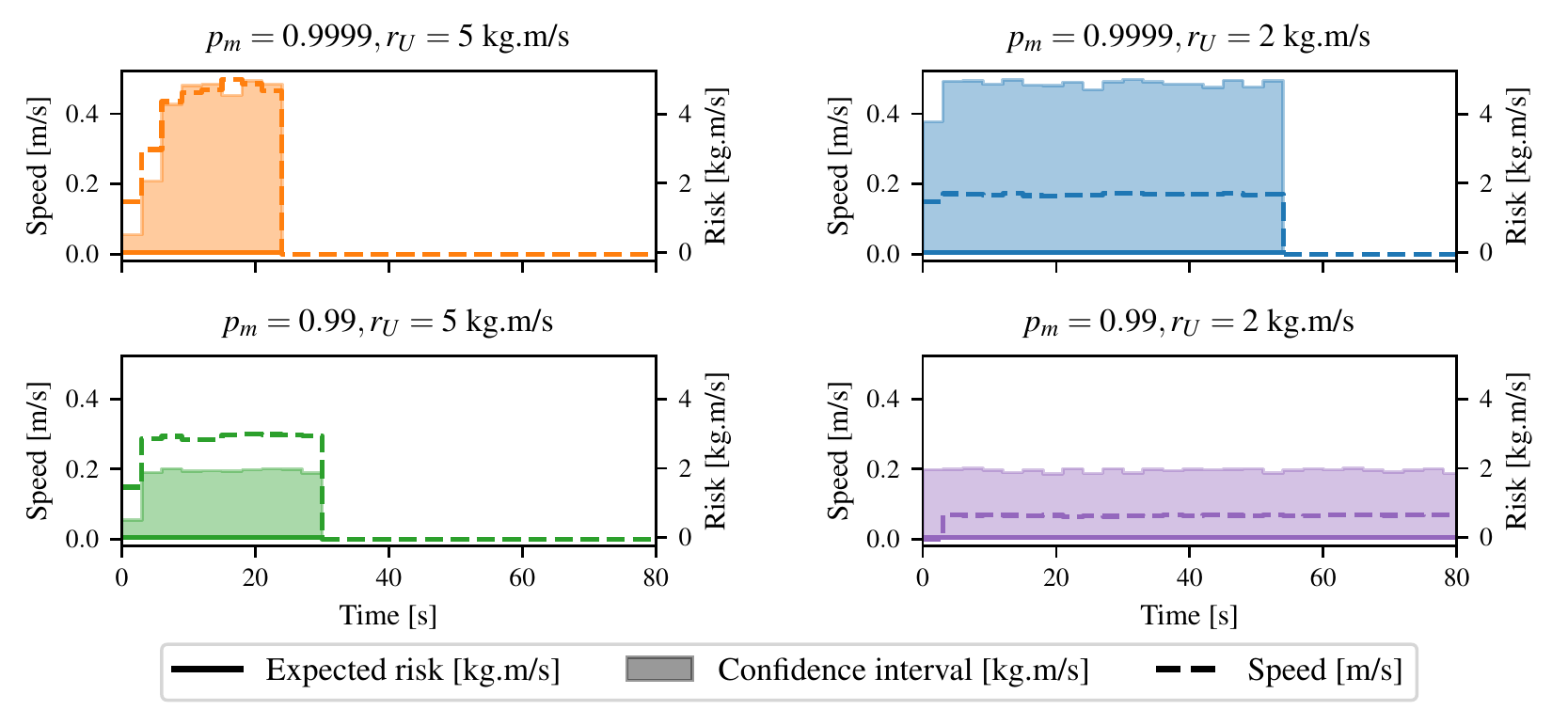}
		\caption{Speed (dashed line) and risk (solid line with a shaded area for its confidence interval) of the chosen paths of the robot going around a tree for different configurations. The~robot was able to navigate at a higher speed when it was confident about the measurements and had a higher upper risk~limit.}
	\label{fig:pathPlanningMetrics_tree}
\end{figure}

Using both algorithms in the same environment allowed the comparison of the behaviors of the robot in a simple case.
Figure~\ref{fig:basicPathPlanning} shows the results of the path planning for the two environment representations. 
For the Lambda Field, the~maximum allowed expected risk was set to $\SI{0}{\kg\m\per\s}$, meaning that the robot must remain clear of any collisions. 
We see that the paths were much alike, and the robot effectively avoided the obstacles in both cases.
It can be seen for the Lambda Field that the robot path crossed some cells where the lambda was not null, which would lead to a collision.
However, as~the Lambda Field was computed in real time, the~lidar measured collisions in this cell after the robot crossed it.
The lidar beams could indeed go through the grass or returned a collision depending on the position of the robot.

The same experiment was conducted using different parameters for the Lambda Field.
Figure~\ref{fig:pathPlanningMetrics_tree} shows the resulting speed of the robot for different configurations of parameters.
While the robot had to expect no risk on its path, it was first allowed to have an upper risk at $\SI{5}{\kg\m\per\s}$, meaning that we were sure at 95\% that any unexpected collision had an expected risk below $\SI{5}{\kg\m\per\s}$.
The robot quickly reached its maximal speed with full acceleration while keeping the upper risk below the threshold.
Under the same configuration, the~robot had to reach the goal while keeping the upper risk below $\SI{2}{\kg\m\per\s}$.
As the upper risk was smaller, the~robot had to reduce its speed.
The robot had the same type of reaction if its confidence in the lidar sensor decreased.
In the third experiment, the~probability of a correct `miss' measurement $p^m$ was decreased from $0.9999$ to $0.99$.
The direct implication was that the confidence interval broadened, forcing the robot again to decrease its speed.
Then, the~robot had a poor confidence in the lidar, as well as a small upper risk, leading to a very slow traversal speed.
Thereupon, the~Lambda Field allowed classical path planning in the same fashion as in~\cite{Heiden2017}.
Our framework also regulated the speed of the robot to cope with the allowed risk~level.

\subsection{Going Through Tall~Grass}
	After performing simple path planning, we show that the Lambda Fields allow the robot to navigate in unstructured environments. %
	As shown in Figure~\ref{fig:pathPlanningTallGrass}a, the~robot had to reach a goal that was behind tall grass.
	This kind of environment leads to very noisy maps, which can hinder the robot's displacements if looking at the probability of~collision.
	We here show that according to the risk the robot is willing to take, it chooses to either go through the tall grass or tries to find a path around it.
	This kind of behavior is impossible to have when only looking at the probability of collision.
	Indeed, the~robot is sure to collide with the grass.
	The probability of collision is high, but the collision caused by the grass is harmless, leading to a very small risk.
	Using a camera, the~robot knows that the obstacles in front of it are tall grass.
	For any other zone, the~mass is set to the worst case (i.e., $\mathbb{P}(m_i=\infty)=1$), as any other prior may lead to underestimating the risk.
	We assumed that tall grass has a 95\% chance of having a null mass and a 5\% chance of having an infinite mass.
	This probability models the possibility that tall grass can hide very dense obstacles, such as rocks or tree~trunks.

\end{paracol}
\nointerlineskip
\begin{figure}[H] 
\widefigure
		\begin{tabular}{ccc}
\includegraphics[width=.3\linewidth]{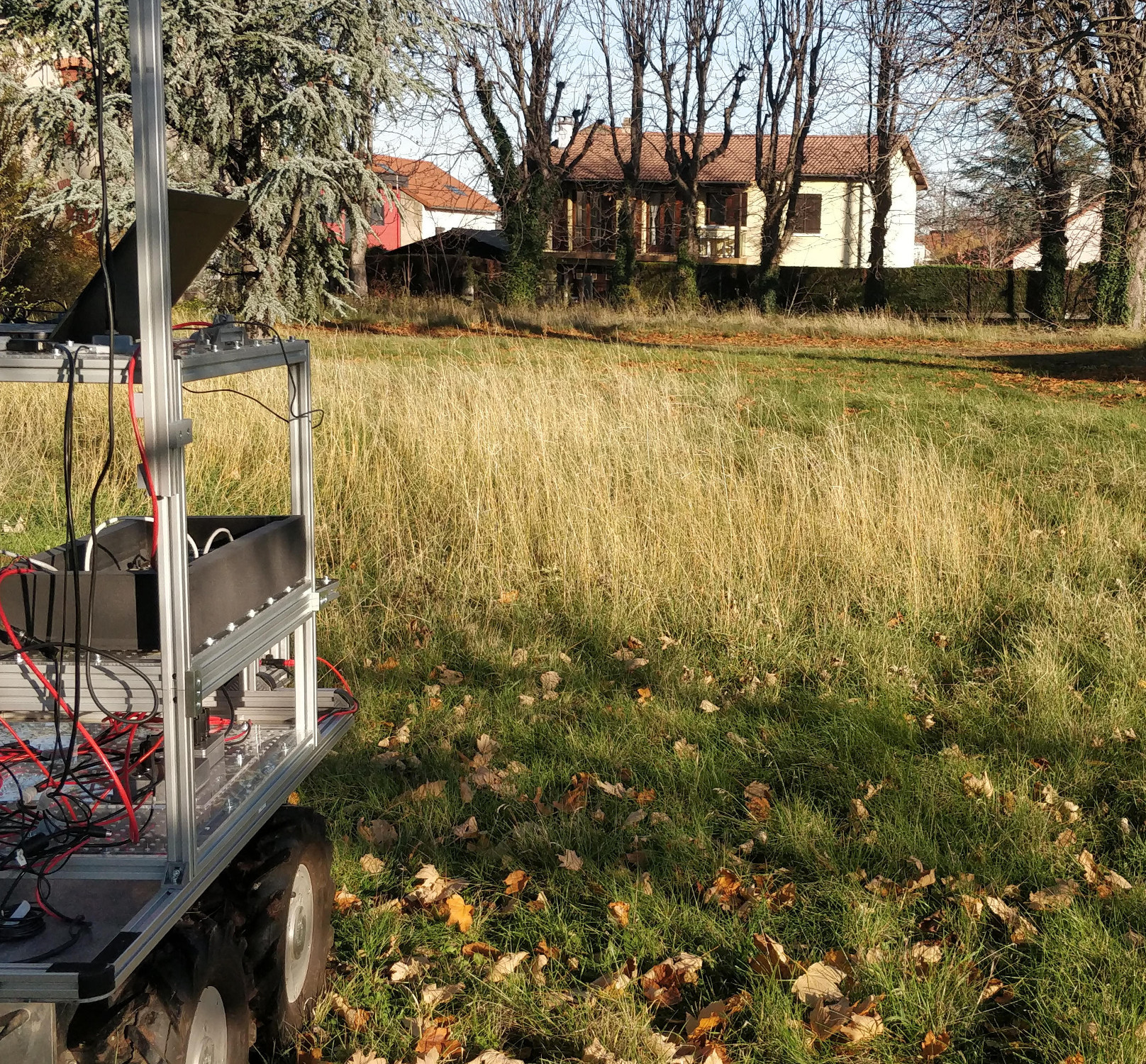}&   \includegraphics[width=.3\linewidth]{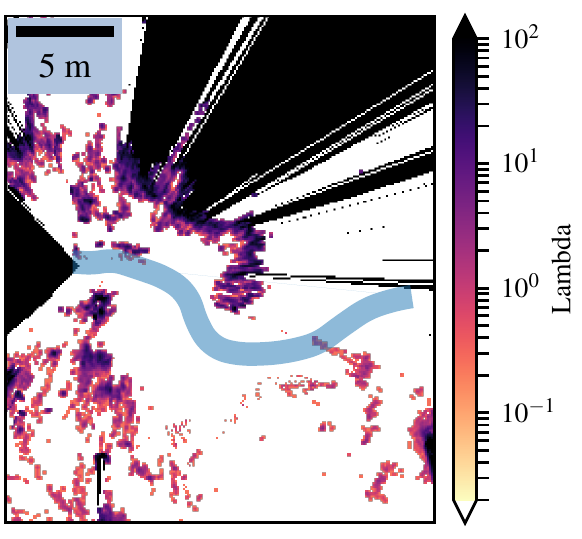} &  \includegraphics[width=.3\linewidth]{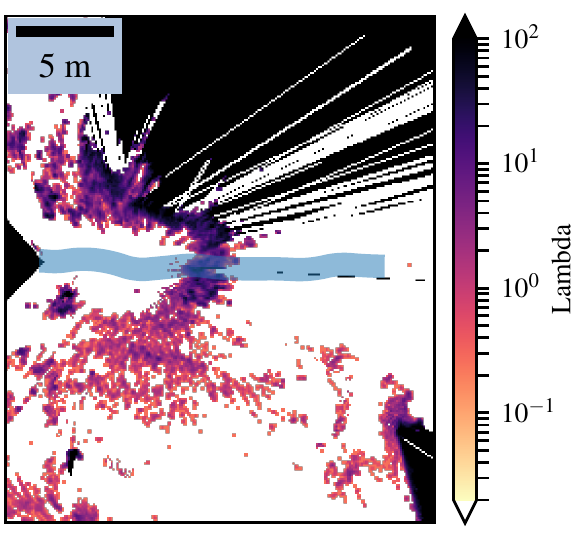}\\
({\bf a})& 
({\bf b})&
({\bf c})\\
	\end{tabular}
		\caption{The robot had to reach a goal behind the tall grass. 
                  (\textbf{a}) Picture of the environment.
                (\textbf{b}) Lambda Field with the path the robot took  in blue, where the robot was instructed to take absolutely no risks. 
              (\textbf{c}) Lambda Field with the path the robot took  in blue, where the robot was allowed to take some~risks.}
		\label{fig:pathPlanningTallGrass}
	\end{figure}
\begin{paracol}{2}
\switchcolumn

	Two cases were analyzed: In the first one, the~robot had to take no risks, meaning that for every path the robot took, the~expectation of the risk had to be zero. Hence, the~robot chose to go around the tall grass.
	In the second case, the~robot was allowed to take some risks to reach its goal and went through the tall grass. 
	Figure~\ref{fig:pathPlanningTallGrass} shows the resulting Lambda Fields for these two different robot configurations.
	In the first case, since the tall grass had a non-zero probability of having a mass that would lead to a harmful collision, the~robot chose to go around the tall grass to reach the goal.
	Once again, the~cells with a lambda higher than zero on the path of the robot were  updated after the robot went through them. At~the time that the robot crossed these cells, the lambdas were null.
	Figure~\ref{fig:pathPlanningMetrics_goingAround} shows the speed as well as the risk taken by the~robot.
	\begin{figure}[H] 
			
			\includegraphics[width=.95\linewidth]{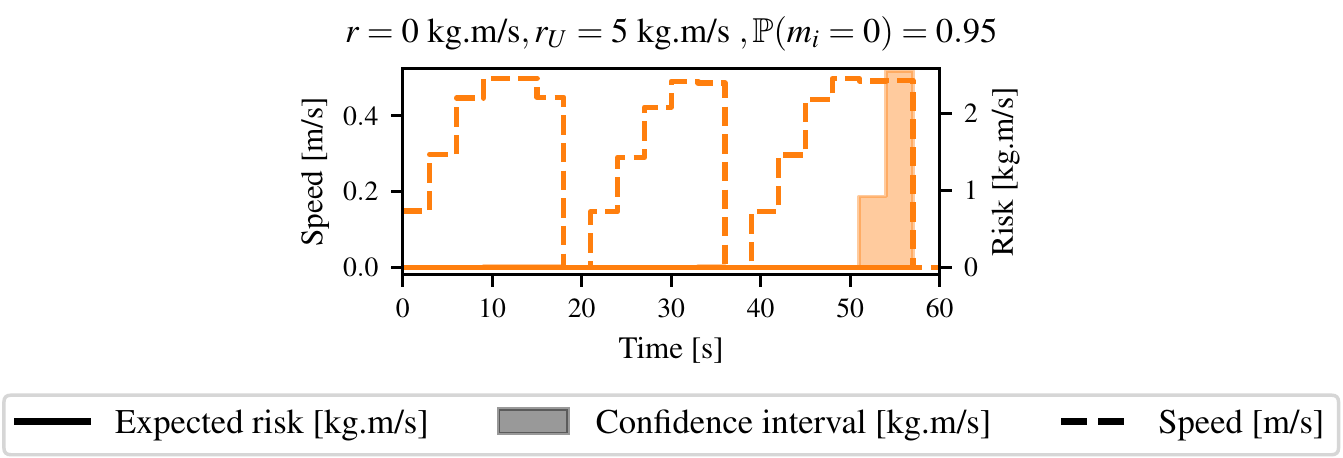}
		\caption{Speed (dashed line) and risk (solid line with a shaded area for its confidence interval) of the chosen paths of the robot when it chose to go around the tall grass. The~numerous stops of the robot were due to the very random detection of the tall~grass.}
		\label{fig:pathPlanningMetrics_goingAround}
	\end{figure}
		
	The robot first crossed a zone where the mass was supposed to be low, leading to a very narrow confidence interval.
	The confidence interval grew quickly as the robot went out of the low-mass zone.
	The robot also stopped several times during the traversal.
	Indeed, a~lot of grass hindered its movements, as the detection of the grass was very irregular.
	Because of the maximum deceleration of the robot, no paths were below the maximum risk allowed.
	The robot then had no choice but to completely stop to be able to plan with low-speed~commands.

	In the second case, the~robot was allowed to take some risks and chose to go through the tall grass to reach the goal.
	The differences in the lambdas between the two maps came from the fact that depending on the robot position, the~lidar beams could go through the grass or return a collision. 
	Furthermore, there was a lot of wind during the experiments, leading to an accentuation of the noise of the overall map.
	For this case, different configurations of risk were analyzed.
	Figure~\ref{fig:pathPlanningMetrics_goingThrough} shows the speed as well as the risk taken by the robot. %
	In the first configuration, the~robot entered the grass at $t\approx\SI{12}{\s}$.
	It was allowed to have an expected risk of $\SI{0.1}{\kg\m\per\s}$ and an upper risk of $\SI{5}{\kg\m\per\s}$.
	The grass had a 5\% chance of having an infinite mass.
	The robot stopped at $t\approx\SI{18}{\s}$ as it was about to enter a denser zone, meaning a zone with a higher collision probability.
	All the high-speed commands led to  too high of a risk, and the robot had to completely stop.
	During the traversal of the tall grass, the~speed of the robot was maintained at a low value, as the grass might have been hiding an obstacle.
	As the robot went out of the grass at $t\approx\SI{36}{\s}$, it increased its speed to its maximum, since a collision was more unlikely to happen.
	The same experiment was conducted, but this time, the grass zone had a probability of 99\% of having a null mass.
	The robot stopped in the same place as the first time, but~increased its speed faster as it was more sure that the collisions were harmless.
	By doing so, it reached the goal more quickly.
	The third time, the~robot was sure that there were no obstacles in the grass. %
	Hence, it crossed the environment at full speed, since any collisions were harmless, even though the maximum risk allowed was null.
	A very specific event could appear: Sometimes, the~expected risk was outside the confidence interval.
	This can be seen for $t=\SI{25}{\s}$ in the first graph in Figure~\ref{fig:pathPlanningMetrics_goingThrough}.
	Computing the risk with lower lambdas could indeed lead to a higher risk in very specific conditions.
	In our case, the~robot computed the risk for a path going out of the area, where the mass of the obstacles was classified as low (i.e., the tall grass) by the camera.
	By doing so, any collision happening outside this zone would have a higher expected force of collision.
	It was then considered less risky to collide inside the area of low mass; lowering the lambdas led to a higher chance of colliding outside, as the robot had a lower chance of being stopped inside the zone, leading to a higher~risk.
	\begin{figure}[H] 
			\includegraphics[width=.9\linewidth]{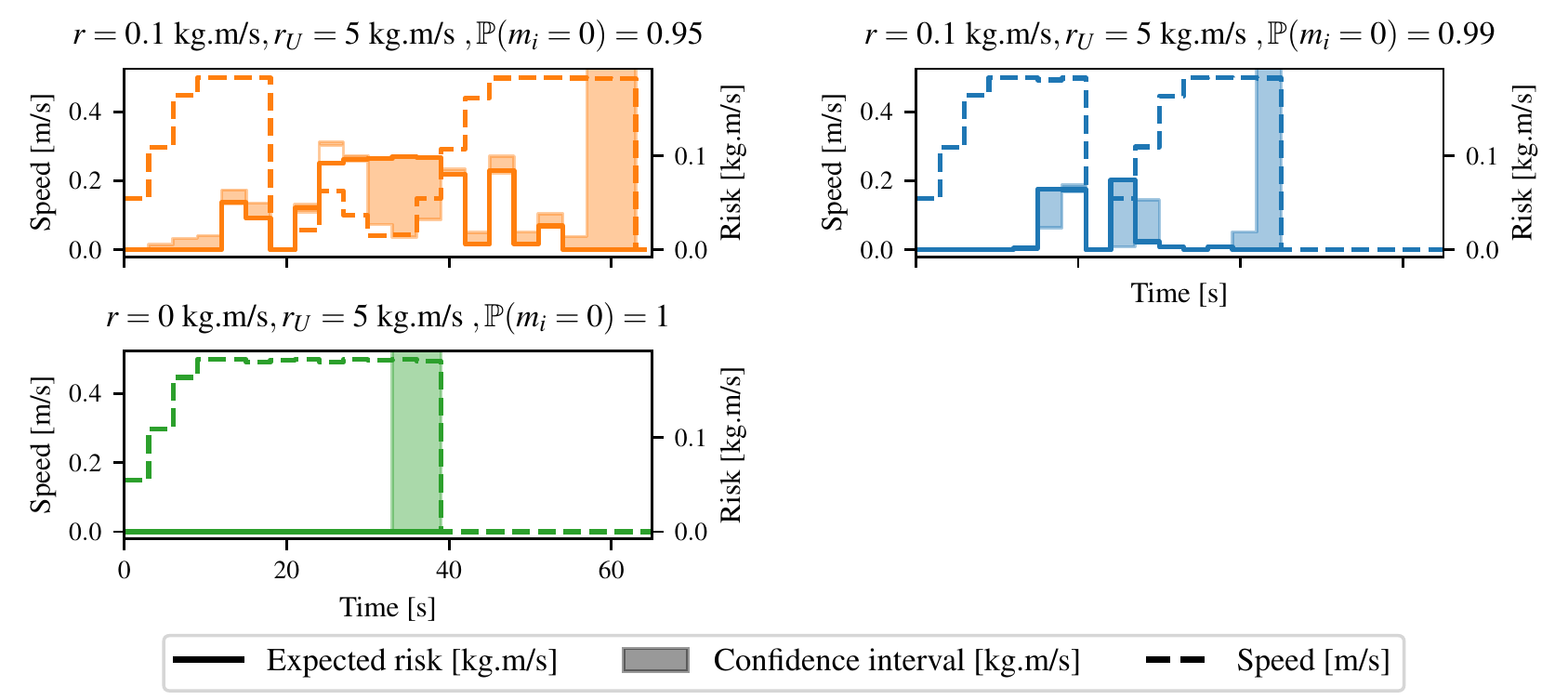}
		\caption{Speed (dashed line) and risk (solid line with a shaded area for its confidence interval) of the chosen paths of the robot when it chose to go through the tall grass for~different configurations. 
		The more certain the robot was that there were no obstacles in the grass, the~faster it reached its goal.}
		\label{fig:pathPlanningMetrics_goingThrough}
	\end{figure}

\section{Discussion}
{The framework was implemented on a standard-grade CPU and ran at more than \mbox{\SI{10}{\Hz}}.
Evidently, the~smaller the tessellation, the~more slowly the framework will operate.
If much larger maps or a finer path planner are needed, the~whole framework can easily run in parallel on a GPU.
Each cell can be updated independently, whereas all of the potential paths can be assessed at the same time.
However, we found that for standard applications, a~cell size below \mbox{\SI[parse-numbers=false]{10\times 10}{\cm}} is not necessary and does not yield better results, which is mainly due to the sensor's noise, and~evaluating \mbox{\num{300}} commands at each iteration is enough for a smooth~navigation.
}

As mentioned before, the~theory of the Lambda Field can be seen as a generalization of the framework of~\cite{Heiden2017}.
Under this consideration, it is possible to convert a Lambda Field into a Bayesian occupancy grid and vice~versa using Equation (\ref{eq:relationReachabilityLambda}).
In addition to adding meaning to the equations of~\cite{Heiden2017}, our framework allows the computation of expectation of a risk.
This is possible because the Lambda Field possesses a probability density function.
Furthermore, the~theory of the Poisson point process was already  used by~\cite{Eggert2014} for known obstacles.
The Lambda Field can be seen as the transposition of their work for occupancy~grids.

\textls[-15]{One of the major drawbacks of the Lambda Field is the assumption  in \mbox{Equation (\ref{eq:mapping_approx})}, which is that every cell in the error region of the range sensor carries the same information.
Using such an approximation indeed leads to inflation of the obstacles, meaning that some narrow corridors through which the robot could go  become impracticable.
The modeling of the sensor can also be discussed: For practical reasons, the~sensor is assumed to have a deterministic error region where the collision is sure to have happened.
Selecting too small of an error region would lead to augmenting the probability of wrong measurements, thus increasing the confidence interval of the lambdas and, hence, decreasing the speed of the robot.
Inversely, taking too big of an error region leads the map to inflate the obstacles, hence decreasing the space in which the robot can evolve.  
However, in~the case of lidar sensors, their precision is such that their error region often reduces to a low number or even a single cell, meaning that almost no inflation occurs.
This can be seen in Figure~\ref{fig:wireFence} or Figure~\ref{fig:compareMaps}, for instance, where the structured obstacles do not appear bigger on the Lambda Field than in the Bayesian occupancy grid.
In the case where the range sensor has a bigger error zone, Appendix \ref{appendix:probabilisticErrorRegion} gives a way to reduce the inflation by using a standard probabilistic sensor~model.}

The computation of the confidence intervals can also yield  deeper discussions.
Indeed, an~empty cell close to an obstacle can be considered to have greater chances of failing the reading and returning a `hit' measurement.
Although this problem is already managed by the error zone, taking into account that cells close to the error zone have a greater chance to be misread can lead to more precise maps.
This would lead to a lower  upper bound of the lambdas in a large empty zone, thereby increasing the traversal speed (i.e., efficiency) of the robot.
As this article only dealt with constant false positive and negative ratios, future works will investigate a deeper use of confidence intervals in harsher environments, such as snowstorms, where many faulty `hit' measurements coming from the snow can hinder the robot~displacements.

So far, only the lidar measurements are used to estimate the lambdas, and a camera is used for the mass estimation.
However, in~a more cooperative approach, the~robot can also assess the hazardousness of the environment while navigating in it.
For instance, if~the robot goes safely through a zone where the probability of causing a harmful collision $p_{si}$ is not null, the~map can be updated accordingly and the probability of harmful collisions can drop.
Note that adding such information needs to be done carefully, as~informing that a zone is safe for a robot does not mean that another robot with a different mass and mechanical configuration will  also safely  cross.
Furthermore, as~the framework creates maps centered on the robot, there is currently no tool for matching temporally distant observations (such as loop closures).
In a case where the construction of large-scale maps is necessary, an~external Simultaneous Localization and Mapping (SLAM) algorithm, such as the Iterative Closest Point (ICP), can provide a corrected~localization.

In addition, some issues can appear while estimating the risk with our framework for global path-planning algorithms.
Indeed, it may be harder to understand the metrics.
The longer the path, the~higher the risk will be.
As this behavior seems intuitive, it leads to several questions.
For two paths with the same risk but a different length, are we willing to take the two paths with the same confidence?
 Should we weigh the risk by the length of the path, meaning that we are willing to take more risk for longer paths?
We chose to understand the risk as the risk of a given command.
Indeed, we believe that as humans, we assess the risk of every step we make without thinking about the length of the~path.

Furthermore, the~expectation of the risk is not suited to modeling the `long-tail' of the Gaussian, i.e., the low-probability events that can happen.
The faulty measurements of the lidar are handled with the confidence intervals of the lambdas, but other metrics may better estimate the `long-tail'.
The Conditional Value at Risk presented by~\cite{Majumdar2017} explicitly measures the risk of the `long-tail'.
It can then be a better indicator of the risk when a low-probability, high-risk situation arises; for~example, when a high-mass obstacle hides in the~grass.

\section{Conclusions}
	In this article, we presented  a novel representation of the occupancy information of the environment, which is called the Lambda Field.
	We first derived a way to construct the map, as~well as confidence intervals over these values.
	This representation allows the computation of expectations over a path, giving a natural way to assess different types of risks.
	The Lambda Field is very similar to the Bayesian occupancy grid for mapping, with~the only notable mapping difference being that the Lambda Field better stores unstructured obstacles, such as bushes, tall grass, or wire fences.
	In addition, the~Lambda Field provides the computation of a generic risk that depends on the~application.

	In the case of unmanned ground vehicles, we chose to represent the risk as the force of collision.
	In contrast to risk metrics defined on Bayesian occupancy grids, our risk possesses a physical meaning.
	We were able to control the level of risk that the robot could take over its planning, allowing behaviors that are impossible with classical path-planning representations of the environments.
	The robot was indeed allowed to cross low-mass occupied areas, such as tall grass, as long as the risk level was low enough.
	Therefore, the~Lambda Field provides a framework that regulates the path as well as the speed of the robot, ensuring the robot's~safety.%

	Future works will investigate the use of other metrics than the expectation of the risk while testing the framework in more adverse environments, such as snowstorms or deep woods.
	Proprioceptive information of the robot will be included to fill the map more accurately.
	Finally, the~framework will be improved to take into account dynamic obstacles and assess risks in urban-like environments.
	Using the risk function instead of the standard probability of collision will lead to more informed decisions in case of~danger.

\vspace{6pt} 

\authorcontributions{
Conceptualization, J.L.; 
methodology, J.L.; 
software, J.L. and L.M.; 
validation, J.L., A.K., and L.M.; 
formal analysis, J.L.; 
writing---original draft preparation, J.L. and A.K.; 
writing---review and editing, R.A., F.P., R.C., and C.D.; 
visualization, J.L. and F.P.; 
supervision, R.A., F.P., R.C., and C.D.; 
funding acquisition, R.A. All authors have read and agreed to the published version of the~manuscript.
}

\funding{
This work was sponsored by a public grant overseen by the French National Research Agency as part of the “Investissements d’Avenir” through the IMobS3 Laboratory of Excellence (ANR-10-LABX-0016), the~IDEX-ISITE initiative CAP 20-25 (ANR-16-IDEX-0001), and the RobotEx Equipment of Excellence (ANR-10-EQPX-0044).
}

\acknowledgments{
  We thank Elie Randriamiarintsoa for his help with the implementation of the image segmentation framework.
  We also thank Simon Vilmin for his advice and the insightful discussions we had about the comparison with related~works.
}
\conflictsofinterest{The authors declare no conflict of~interest.} 

\appendixtitles{yes} %
\appendixstart
\appendix
\appendix
\section{Heterogeneous Error~Regions}
\label{appendix:heterogeneousErrRegions}
	In the case that the error regions $\mathcal{E}_k$ have a different size for each \ac{lidar} beam $b_k$, we need to further approximate the derivative of the log-likelihood.
	Under the same assumption that the $h_i$ error regions $\mathcal{E}_k$ containing the cell $c_i$ are small, we have%
\begin{equation}
		\begin{aligned}
			\frac{\partial \mathcal{L}(X|\lambda)}{\partial \lambda_i} %
	&= -m_c\cdot\area + \sum_{k=0}^{h_i-1}\frac{\area}{\exp(e_k\lambda_i)-1} \\
	&\approx -m_c\cdot\area + \sum_{k=0}^{h_i-1}\frac{\area}{e_k\lambda_i} ,\\
		\end{aligned}
	\end{equation}
	leading to
\begin{equation}
		\lambda_i = \frac{1}{m_i}\sum_{k=0}^{h_i-1}\frac{1}{e_k} .
		\label{eq:lambdaDifEk}
	\end{equation}
	
	This approximation over-estimates the lambdas compared to Equation (\ref{eq:lambda}).
	Indeed, in~the special case where all the $\mathcal{E}_k$ have the same area $e$, the~computed lambdas from Equation (\ref{eq:lambdaDifEk}) are
\begin{equation}
		\lambda_i = \frac{1}{e}\frac{h_i}{m_i} .
	\end{equation}
	
	As $\forall x\in\mathbb{R}_{\geq 0},\, x\geq \ln(1+x)$, we will always over-estimate the lambdas using \mbox{Equation (\ref{eq:lambdaDifEk})}.
	This is the desired behavior, as under-estimating the lambdas would lead to under-estimate the~risk.
\section{Proof of Equation (\ref{eq:expected_coll_masses})}
\label{appendix:proofExpCollMass}
	We here prove Equation (\ref{eq:expected_coll_masses}).
	We have two random variables: the area at which the robot collides $A$ and the mass $M$ of the cell where the collision happened, of~marginal probability density functions $f(\cdot)$ and $f^m(\cdot)$.
        Note that we do not have direct access to $f^m(\cdot)$, but only $f^m_i(\cdot) = f^m(\cdot | \area i)$, the~probability density function of the mass given the crossed area, thus given the cell that the robot is currently crossing.
	Under the assumption that the risk $r(\cdot)$ is constant inside each cell, the~expectation of the function $r(A,M)$ is
\begin{equation}
		\begin{split}
			&\mathbb{E}\left[r(A,M)\right] = %
			\int_0^{N\area}\int_0^\infty r_m(a,m)f(a) f^m(m|a)\dif m \dif a \\
			=& \sum_{i=0}^{N-1} \int_{i\area}^{(i+1)\area}f(a) \int_0^\infty r_m(a,m) f^m_i(m)\dif m \dif a \\
			=& \sum_{i=0}^{N-1} \left[\int_{i\area}^{(i+1)\area}f(a)\dif a\right]\left[ \int_0^\infty r_m(\area i,m) f^m_i(m)\dif m\right] \\
			=& \sum_{i=0}^{N-1} K_i\int_0^\infty r_m(\area i,m_i)f^m_i(m_i)\dif m \\
			=& \sum_{i=0}^{N-1} K_i\sum_{k=0}^\infty \alpha_{ik} r_m(\area i, k\Delta_m),
		\end{split}
	\end{equation}
	for a path $\mathcal{P}$ going through the cells $\{c_i\}_{0:N-1}$, and~\begin{equation}
          K_i = \expp{-\Lambda_m(\{c_j\}_{0:i-1})}\left[1-\expp{-\Lambda_m(\{c_i\})}\right].
        \end{equation}

\section{Probabilistic Error~Region}
	\label{appendix:probabilisticErrorRegion}
	If the range sensor has a large error zone, the~inflation of the obstacles may become problematic.
	Therefore, we give here a way to estimate the Lambda Field using a probabilistic error region.
	Let the error region $\mathcal{E}: \mathbb{R} \rightarrow \mathcal{P}(\mathbb{R}^2)$ be an application that takes a parameter and returns a subspace of $\mathbb{R}^2$. 
	One example is the application that gives from the radius $r$ the ball centered on the lidar measurement $\bm{x_l}\in\mathbb{R}^2$: $\{\bm{x}\in\mathbb{R}^2 , |\bm{x}-\bm{x_l}|\leq r\}$.
	Furthermore, let $\sigma$ be a random variable of probability density function $f_\sigma(\cdot)$ and a Lambda Field $\lambda:\mathbb{R}^2\rightarrow\mathbb{R}_{\geq 0}$. 
	Under these considerations, the~expectation of the intensity of the error zone $\mathcal{E}(\sigma)$ is
\begin{equation}
	\begin{aligned}
		\mathbb{E}\left[ \int_{\mathcal{E}(\sigma)} \lambda(\bm{x})\dif \bm{x}  \right]%
		&= \int_\mathbb{R} f_\sigma(s) \int_{\mathcal{E}(s)} \lambda(\bm{x}) \dif \bm{x} \dif s \\
		&= \int_\mathbb{R} f_\sigma(s) \int_{\mathbb{R}^2} \lambda(\bm{x})\cdot\mathds{1}_{\mathcal{E}(s)}(\bm{x}) \dif \bm{x} \dif s.
	\end{aligned}
	\end{equation}
	
	Under the assumption that the expectation of the intensity of the error zone is finite, we can switch the integration order using Fubini's theorem and find a more convenient~form:
\begin{equation}
	\begin{aligned}
		\mathbb{E}\left[ \int_{\mathcal{E}(\sigma)} \lambda(\bm{x})\dif \bm{x}  \right]%
		&= \int_{\mathbb{R}^2} \lambda(\bm{x}) \int_{\mathbb{R}} f_\sigma(s)\cdot\mathds{1}_{\mathcal{E}(s)}(\bm{x}) \dif s \dif \bm{x} \\
		&= \int_{\mathbb{R}^2} \lambda(\bm{x}) \,\mathbb{P}(\bm{x}\in \mathcal{E}(\sigma)) \dif \bm{x},
	\end{aligned}
	\end{equation}
	where $\mathds{1}_X$ is the identity operator, i.e.,~$\mathds{1}_X(\bm{x}) = 1$ if $\bm{x}\in X$ and $0$ otherwise.
        Using this expectation as the new intensity function $\Lambda(\mathcal{C})$, tessellating the field and putting it back into Equation (\ref{eq:logLikelihood}), the~lambdas are now estimated using the same counters $h_i$ and $m_i$, which now represent, respectively, the sum of the probabilities of being in the error zone and the sum of the probabilities of not being in the error zone of each lidar measurement.
	One can note that in the case where the error region is known, meaning that $\mathbb{P}(x\in \mathcal{E}(\sigma))$ is either equal to zero or one, we fall back on the previously derived equations as $h_i$ and $m_i$ regain their function of counting the number of times the cell has been in the error~region or not.

\vfill	

\end{paracol}
\reftitle{References}

\end{document}